\documentclass[10pt,twocolumn,letterpaper]{article}

\usepackage[cvprpagenumbers]{cvpr}      
\usepackage{times}
\usepackage{epsfig}
\usepackage{graphicx}
\usepackage{amsmath}
\usepackage{amssymb}
\usepackage{soul}

\usepackage{algorithm}
\usepackage{algorithmic}

\usepackage{bbold}
\usepackage{bm}
\usepackage{pgfplots}
\pgfplotsset{width=0.8\textwidth, compat=1.5}
\usepackage{enumerate}
\usepackage{enumitem}
\usepackage{caption}
\usepackage{colortbl}
\usepackage{multirow}
\usepackage{pgf-pie}
\usetikzlibrary{shadows}
\usepackage{mathtools}
\usepackage{booktabs}
\usepackage{amssymb}

\usepackage{newfloat}
\usepackage{listings}
\usepackage{setspace}
\usepackage{utfsym}

\definecolor{cvprblue}{rgb}{0.21,0.49,0.74}
\usepackage[pagebackref,breaklinks,colorlinks,citecolor=cvprblue]{hyperref}

\def\paperID{72} 
\def\confName{3DV\xspace}
\def\confYear{2024\xspace}

\title{Enhancing Generalizability of Representation Learning for Data-Efficient 3D Scene Understanding}

\author{Yunsong Wang\textsuperscript{1} \quad Na Zhao\textsuperscript{2}\thanks{This work was done when Na Zhao was a visitor at NUS.} \quad Gim Hee Lee\textsuperscript{1}\\
\textsuperscript{1}National University of Singapore\\
\textsuperscript{2}Singapore University of Technology and Design\\
{\tt\small yunsong@comp.nus.edu.sg, na\_zhao@sutd.edu.sg, gimhee.lee@comp.nus.edu.sg}
}

\begin{document}
\maketitle

\begin{abstract}
    The field of self-supervised 3D representation learning has emerged as a promising solution to alleviate the challenge presented by the scarcity of extensive, well-annotated datasets. However, it continues to be hindered by the lack of diverse, large-scale, real-world 3D scene datasets for source data. To address this shortfall, we propose Generalizable Representation Learning (GRL), where we devise a generative Bayesian network to produce diverse synthetic scenes with real-world patterns, and conduct pre-training with a joint objective.
    By jointly learning a coarse-to-fine contrastive learning task and an occlusion-aware reconstruction task, the model is primed with transferable, geometry-informed representations.
    Post pre-training on synthetic data, the acquired knowledge of the model can be seamlessly transferred to two principal downstream tasks associated with 3D scene understanding, namely 3D object detection and 3D semantic segmentation, using real-world benchmark datasets. A thorough series of experiments robustly display our method's consistent superiority over existing state-of-the-art pre-training approaches.
\end{abstract}

\section{Introduction}
3D scene understanding that perceives and understands the three-dimensional world is crucial for many real-world applications such as self-driving cars, domestic robots, augmented and virtual reality devices, \textit{etc}. Over the past few years, deep learning-based 3D scene understanding tasks such as 3D semantic segmentation~\cite{pointnet, pointnet++, kpconv, sr-unet} and 3D object detection~\cite{frustum, votenet, h3dnet, group-free} have attracted growing interest and achieved remarkable progress. Nonetheless, the success of these deep models heavily relies on large-scale well-annotated datasets. The existing 3D scene-level datasets~\cite{s3dis, scannet, sunrgbd} are highly limited in terms of size 
due to the much higher cost of annotations in 3D tasks comparing to those in 2D tasks.

\begin{figure}[ht]
    \centering
    \includegraphics[width=1\linewidth]{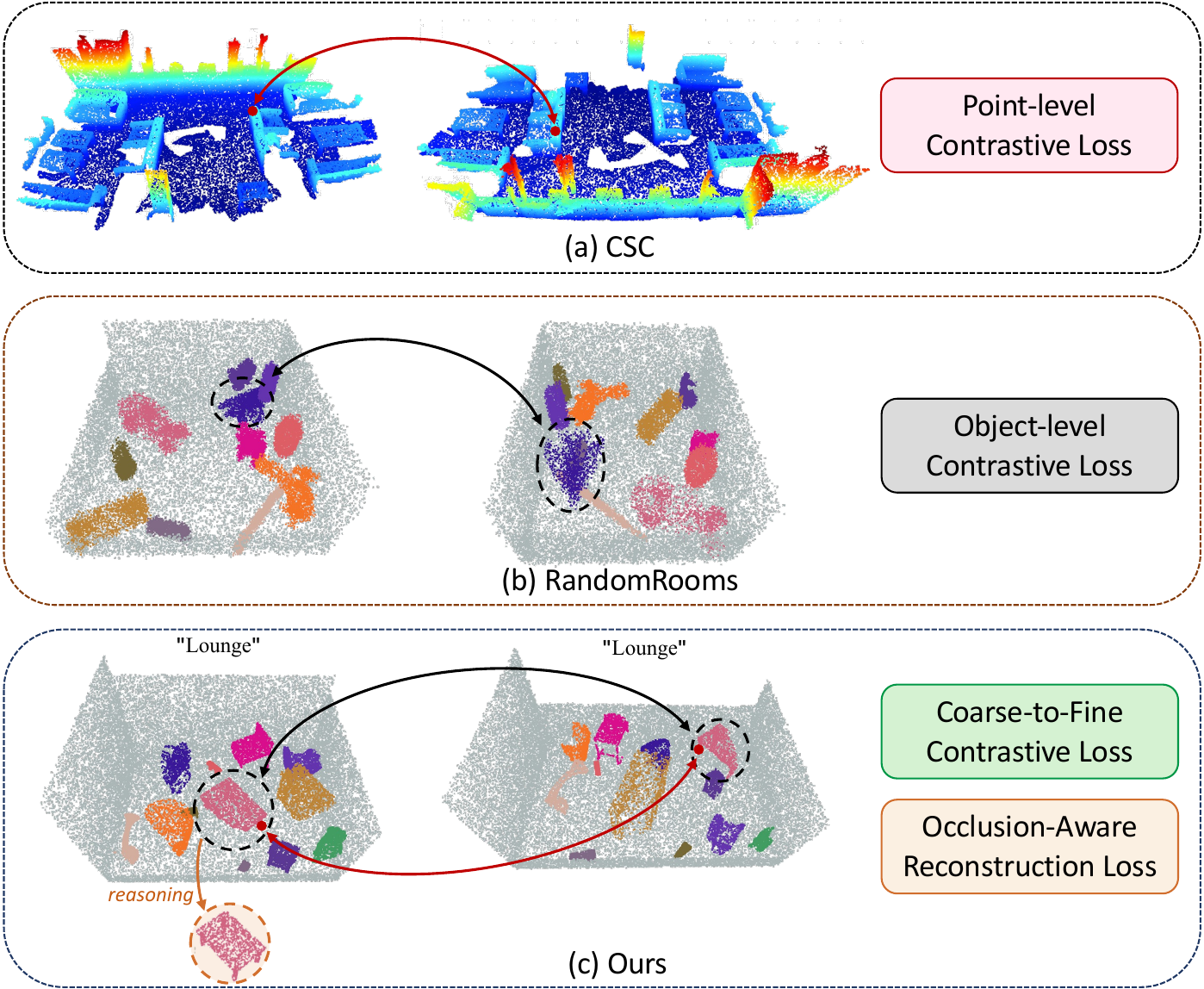}
    \caption{\small{\textbf{Comparison of representation learning methodologies.} We compare our overall framework with CSC \cite{partition} and RandomRooms \cite{randomrooms}.}}
    \vspace{-0.1in}
    \label{teaser}
\end{figure}


In view of the scarcity of large-scale well-annotated 3D data, self-supervised representation learning that can train a transferable representation learner from unlabeled data becomes an appealing solution to boost the performance. 
Recently, a number of self-supervised learning works have been proposed, but most of them require specialized pre-training data that can be difficult to satisfied. For example, both PointContrast \cite{pointcontrast} and CSC \cite{partition} require point-wise correspondences across different views to form positive sample pairs for point-level contrastive learning, which require scene-level point registration and can limit their scalability of pre-training data.
The commonly used pre-training dataset, ScanNet \cite{scannet}, is limited to only 1,513 scenes with repetitive rooms, which unavoidably leads to insufficient pre-training and limited capacity for transferring knowledge.


To overcome the absence of large amounts of diverse 3D scenes, RandomRooms \cite{randomrooms} utilizes 3D synthetic object dataset to generate diverse synthesized scenes with random object layouts and performs object-level contrastive learning to enforce the object-level feature consistency. 
Nonetheless, due to their naive iteration over ShapeNet \cite{shapenet} to sample objects and insufficient scene augmentation, their synthesized scenes lack generalizable real-world patterns, including object distributions, scene-object contexts and occlusion. Moreover, their pre-training objective is insufficient to capture the fine-grained object representation. These limitations prevent RandomRooms from carrying out a more effectively transferable representation learning.
In this paper, we propose a Generalizable Representation Learning (GRL) framework, aiming to further enhance the generalizability of the scene-level representation and benefit diverse downstream tasks across datasets and data-efficient settings. 
As depicted in Figure \ref{teaser}, our GRL approach offers several advantages over previous methods. Specifically, for data generation, we devise a simple-yet-effective generative Bayesian network to create large quantities of synthetic scenes with real-world patterns and high diversity. 
For training objective, we equip the pre-trained model with both fine-grained object representation and geometry-reasoning ability, in order to generalize to diverse downstream tasks. We first propose a coarse-to-fine contrastive learning task that learns object-level and point-level representations hierarchically, in order to encode consistent object representations with fine-grained details. Additionally, to account for occlusion and detailed geometries in real-world point clouds, we perform occlusion-aware point cloud reconstruction as a pretext task, encouraging the backbone to preserve the rich geometrical information. The resulting representations obtained through joint learning are capable of capturing generalizable geometry patterns in 3D scenes and can be effectively transferred to various downstream tasks.
Our \textbf{contributions} are summarized as follows: 

\begin{itemize}[leftmargin=2ex, topsep=1pt, itemsep=0.2ex]
    \item We introduce Generalizable Representation Learning (GRL) framework, which comprises of a generative Bayesian network to model scene-object distributions, and a joint learning objective to mine the generalizable geometries using large amounts of synthetic scenes.
    \item Our GRL demonstrates high generalizability, leading to superior boost on variety of tasks and datasets.
    \item 
    We further evaluate GRL under various data-efficient and representation quality settings, where extensive experiments demonstrate our consistent superiority over other existing methods. 
\end{itemize}


\section{Related Work}


\noindent\textbf{3D Deep Learning Backbone Networks.} In recent years, 3D Deep Learning \cite{group-free, morphing, pointconv, pointgroup} is becoming increasingly popular, with point clouds, voxels or meshes as the input formats. To conduct convolution operations on point clouds, multiple previous works \cite{sr-unet, voxel1, voxel2, segcloud} have been focused on processing the voxelized data, and one representative architecture is the SR U-Net \cite{sr-unet}. On the other hand, recently point-cloud based models \cite{pointnet, pointnet++, pointtransformer, so-net} have largely benefited from the convenient input and lower computational cost, where PointNet \cite{pointnet} first leverages deep learning in operating on the unordered point clouds input, and PointNet++ \cite{pointnet++} performs hierarchical sampling and grouping to capture the local patterns that are missing in PointNet. In order to perform pre-training on a more commonly used backbone and benefit from raw point cloud inputs, we use PointNet++ \cite{pointnet++} as our backbone network.

\begin{figure*}[t!]
    \centering
    \includegraphics[width=1\textwidth]{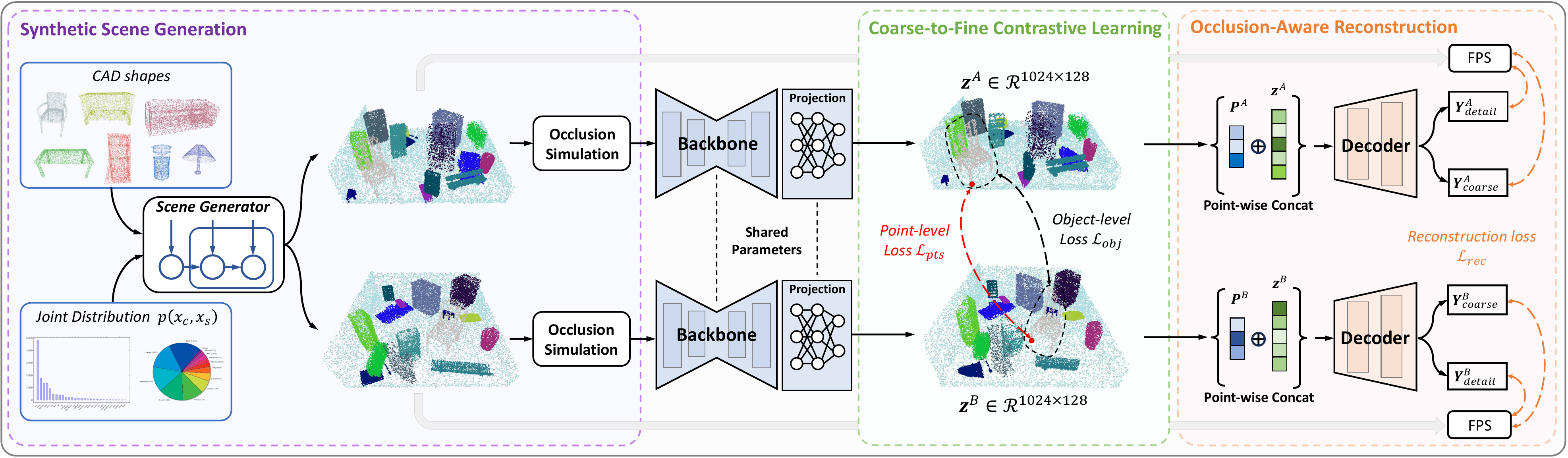}
    \caption{\small{\textbf{Overall structure of pre-training framework.} Our GRL framework devises a generative Bayesian network as the scene generator, whose generated synthetic scenes are augmented with occlusion. Subsequently, we jointly learn a coarse-to-fine contrastive learning task and an occlusion-aware reconstruction task. Note that we show full points of $\bm{z}_A$ and $\bm{z}_B$ for better visualization.}}
        \vspace{-0.1in}
    \label{overall}
\end{figure*}

\vspace{5pt} \noindent\textbf{3D Representation Learning.} Unlike in 2D representation learning \cite{simclr, moco, moco-v2, wang2021instance, caron2021emerging, jia2021scaling}, where the model is typically pre-trained on rich source dataset like ImageNet \cite{imagenet}, 3D representation learning still greatly suffers from the insufficiency of scene-level real-world source datasets. Therefore, most previous works focus on single object level tasks including reconstruction and classification \cite{3Dp1, 3Dp2}. Only recently, motivated by the success of contrastive learning in 2D \cite{bank, simclr}, PointContrast \cite{pointcontrast} proposes to use point-level contrastive learning to pre-train on ScanNetV2 \cite{scannet} and fine-tune on scene-level downstream tasks. Subsequently, CSC \cite{partition} proposes to partition the space for each anchor point and perform point-level contrastive loss in each partition individually. 
Recently, DepthContrast \cite{scenecontrast} proposes to perform scene-level contrastive learning between point clouds and voxel grids without requiring multi-view inputs, and DPCo \cite{dpco} learns the format-invariance between point clouds and depth maps.
Nonetheless, these works still face the scarcity of source dataset. 
To overcome this limitation, RandomRooms \cite{randomrooms} generates large amounts of scenes using synthetic shapes from ShapeNet \cite{shapenet} and applies object-level contrastive learning. However, their naive synthetic scene generation lacks real-world patterns. Moreover, merely applying object-level contrasting fails to capture the fine-grained object representations, leading to limited improvements on downstream tasks like segmentation, which requires explicitly discriminating the object boundaries. To take advantage of the rich synthetic datasets while overcome these weaknesses, we generate synthetic scenes with essential real-world patterns, since it is crucial for the model to be exposed to
the prevalent real-world patterns during pre-training, and we design a joint-objective method to capture generalizable representations.


\vspace{5pt} \noindent\textbf{3D Scene Understanding.} Deep learning for 3D scene understanding has been vastly developed in 3D object detection \cite{frustum, hyperdet3d, h3dnet, imvotenet, od1, od2, od3, od4, od5, votenet}, semantic segmentation \cite{pointnet, pointnet++, segcloud, kpconv, ss1, ss2} and instance segmentation \cite{insseg, SGPN, 3D-SIS, occuseg, is1, is2}. In this work, we choose two of the most representative tasks as our downstream tasks, namely 3D object detection and 3D semantic segmentation, for which we use VoteNet \cite{votenet} and PointNet++ \cite{pointnet++} as the backbone architecture, respectively.




\section{Our Method}
\subsection{Overview}
Figure \ref{overall} illustrates the overall structure of our proposed GRL framework, which includes \textit{synthetic scene generation} and \textit{joint-objective learning}. The synthetic scene generation is based on a simple-yet-effective generative Bayesian network to model the scene-object distributions, and the generated synthetic scenes are further augmented with noise and occlusion. During training, we jointly learn two pretext tasks: 
a coarse-to-fine contrastive learning task and
an occlusion-aware point cloud reconstruction task. 


\subsection{Synthetic Scene Generation}

\begin{figure}[t]
    \centering
    \includegraphics[width=0.3\textwidth]{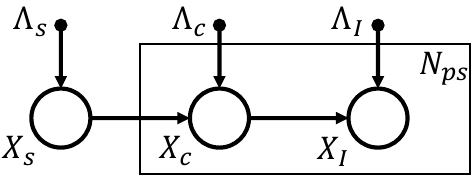}
    \caption{\small{Our scene generator is modeled as a generative Bayesian network. 
    $N_{ps}$ is the number of objects per scene.
    }}
        \vspace{-0.1in}
    \label{mixture}
\end{figure}


\label{rule}
\vspace{5pt} \noindent\textbf{Generative Bayesian Network.}
Unlike the previous work RandomRooms \cite{randomrooms} which naively iterates over ShapeNet \cite{shapenet} to sample objects, we design a 
generative Bayesian network to model scene-object distributions in the real world. 
Figure \ref{mixture} shows our generative Bayesian network.

Let $X_S$, $X_C$ and $X_I$ denote the random variables representing the scene type, object category and object instance, respectively. We model our scene generator as conditional categorical distributions shown in Figure \ref{mixture}, where the joint probability is given by: 
\begin{small}
\begin{align}
    p(X_S, X_C, X_I) = p(X_S)p(X_C \mid X_S) p(X_I \mid X_C).
\end{align}
\end{small}
The prior $p(X_S)$ and conditional probabilities $p(X_C \mid X_S)$ and $p(X_I \mid X_C)$ follow the categorical distributions:
\begin{small}
\begin{align}
    \prod_{k=1}^{N_S} \lambda_{S,k}^{x_{S,k}}, \quad \prod_{k=1}^{N_S} \prod_{l=1}^{N_C} \lambda_{C,kl}^{x_{C,kl}}, \quad \prod_{l=1}^{N_C}\prod_{m=1}^{N_I} \lambda_{I,lm}^{x_{I,lm}}.
\end{align}
\end{small}
$N_S$, $N_C$ and $N_I$ are 
the total number of scene categories, object categories, and instances per object category, respectively.
$x_{S,k}=1$ when $X_S$ takes the $k$-th scene type, and $0$ otherwise. Furthermore, $x_{C,kl} = 1$ when $X_C$ takes the $l$-th object category in the $k$-th scene, and $0$ otherwise. $x_{I,lm}$ is defined in a similar way as $x_{C,kl}$. $\Lambda_S = \{\lambda_{S,1}, \ldots, \lambda_{S,N_S}\}$, $\Lambda_C = \{\lambda_{C,11}, \ldots, \lambda_{C,N_SN_C}\}$ and $\Lambda_I = \{\lambda_{I,11}, \ldots, \lambda_{I,N_CN_I}\}$ are parameters of the categorical distributions, which we learn from the real-world dataset ScanNet~\cite{scannet}. Specifically, taking the maximum log-likelihood of the joint probability over $\Lambda_S$ and $\Lambda_C$, respectively, gives us: 
\begin{small}
\begin{align}
    \lambda_{S,k}^{x_{S,k}} = \frac{N_{S,k}}{\sum_{n=1}^{N_S} N_{S,n}}, \quad \lambda_{C,kl}^{x_{C,kl}} = \frac{N_{C,kl}}{\sum_{n=1}^{N_{C,k}} N_{C,kn}}.
\end{align}
\end{small}
$N_{S,k}$ is the total occurrences of the $k$-th scene and $N_{C,kl}$ is the total occurrences of the $l$-th object category in the $k$-th scene. We can see that $\lambda_{S,k}^{x_{S,k}}$ is simply the normalized occurrence frequency of the $k$-th scene, and $\lambda_{C,kl}^{x_{C,kl}}$ is the normalized occurrence frequency of the $l$-th object category in the $k$-th scene. Since ScanNetV2 does not provide the instance occurrence frequency within specific object categories, we simply set $p(X_I \mid X_C)$ as a uniform distribution.
We then use the learned joint probability distribution to draw samples, where we first draw a sample of the scene category $x_S \sim p(X_S)$, followed by $N_{pc}$ object categories $\{x_C^i\} \sim p(X_C \mid X_S = x_S)$ and the corresponding object instances $x_I^i \sim p(X_I \mid X_C = x_C^i)$. Furthermore, we follow the $\epsilon-greedy$ strategy to switch between uniform and categorical distribution for the object category to induce higher diversity.
Our Generative Bayesian network based synthetic Scene Generation (GBSG) is proposed in a straightforward design to model the real-world scene-object distributions while preserving the inherent diversity of synthetic scenes.
\vspace{5pt} \noindent\textbf{Occlusion Simulation}
To generate synthetic scenes with sufficiently diverse real-world patterns, we perform occlusion simulation on point clouds. 
With hope of incorporating higher diversity, we use the distance-based occlusion simulation similar to the one in PoinTr \cite{pointr}. Specifically, for each scene we randomly pick a viewpoint in the 3D space of the scene, and for each object we randomly remove $0\%$ to $50\%$ furthest points. Such method benefits our framework through its efficiency and occlusion diversity, \textit{i.e.}, each object has random level of incompleteness. 

\subsection{Joint-Objective Learning}


Our generalizable representation learning approach is designed to enable representations to capture the rich geometry information that can be effectively transferred to downstream tasks. Specifically, we propose a coarse-to-fine contrastive learning task to learn consistent fine-grained object representation, and an occlusion-aware reconstruction loss to preserve the detailed geometry information. 

\subsubsection{Coarse-to-Fine Contrastive Learning}

\label{L_obj}
\paragraph{Object-level Constrastive Learning.} Given the two corresponding scenes with occlusion $\Tilde{\bm{R}}_A,\,\Tilde{\bm{R}}_B$, the object-level contrastive loss is calculated based on the the object-level averaged features. Specifically, denoting the backbone network as $g(\cdot)$ and a shared MLP projection as $\varphi_o(\cdot)$, the projection of output features are $\bm{h}^\mathcal{X}=\varphi_o(g(\bm{R}_\mathcal{X}))$, and the instance-level averaged features are $\{\bm{f}_k^\mathcal{X}\}_{k=1}^{K} = \mathcal{A}(\bm{h}^\mathcal{X})$, where $\mathcal{X}\in\{A,B\}$, $K$ is the number of instances per scene, and $\mathcal{A}(\cdot)$ is per-instance average pooling. The object-level contrastive loss is then given by:
\begin{small}
\begin{align}
    \mathcal{L}_{obj} = &-\frac{1}{K}\sum_{k=1}^{K}\text{log}\,\frac{\text{exp}(\bm{f}_k^A\cdot\bm{f}_k^B/\tau)}{\text{exp}(\bm{f}_k^A\cdot\bm{f}_k^B/\tau)+\sum_{\bm{f}\in\mathcal{F}_k}\text{exp}(\bm{f}_k^A\cdot\bm{f}/\tau)} \nonumber \\
    &-\frac{1}{K}\sum_{k=1}^{K}\text{log}\,\frac{\text{exp}(\bm{f}_k^B\cdot\bm{f}_k^A/\tau)}{\text{exp}(\bm{f}_k^B\cdot\bm{f}_k^A/\tau)+\sum_{\bm{f}\in\mathcal{F}_k}\text{exp}(\bm{f}_k^B\cdot\bm{f}/\tau)},
\end{align}
\end{small}
where $\mathcal{F}_k$ is the set of the object features with categories that are different from $\bm{x}_k$ within the batch, $\bm{f}^A_k$ and $\bm{f}^B_k$ is the positive sample pair, $\bm{f}^{\mathcal{X}}_k$ and $\bm{f}\in\mathcal{F}_k$ are the negative pairs, and $\tau$ is the temperature.

Our object-level loss has two additional contributions compared to the loss from RandomRooms \cite{randomrooms}: 
1) \textit{Robustness to occlusion:} The objects in the input scenes have random levels of incompleteness, and thus further encouraging the backbone network to be robust under diverse levels of occlusion. 
2) \textit{Category-aware:} Leveraging the easily accessible object category labels, we only use the features of the inter-class objects as the negative samples 
to focus on discriminating between different object categories. 

\paragraph{Object-aware Point-level Contrastive Learning.} Although the object-level contrastive learning can help represent the object-level geometry and semantic information, it might fail to capture the fine-grained details of object parts, which are important in elevating the competence of the pre-trained model in some downstream task (as shown in Table \ref{ablation}).   
To overcome this problem, we propose an object-aware point-level contrastive loss based on a relaxed point matching scheme. With our point-level contrastive learning, the model can produce consistent representations for specific parts of the same object.

We first introduce how we match points in a pair of scenes. A straightforward way is to sample the exact same points on each object in a pair of scenes and regard them as positive point pairs, which however leads to significant drop on the performance (shown in the supplementary material) due to inconsistent point sampling among the scenes and missing robustness to the randomness of sampling. To circumvent this problem, we propose a relaxed matching scheme based on distances.

\begin{figure}[t]
    \centering
    \includegraphics[width=0.3\textwidth]{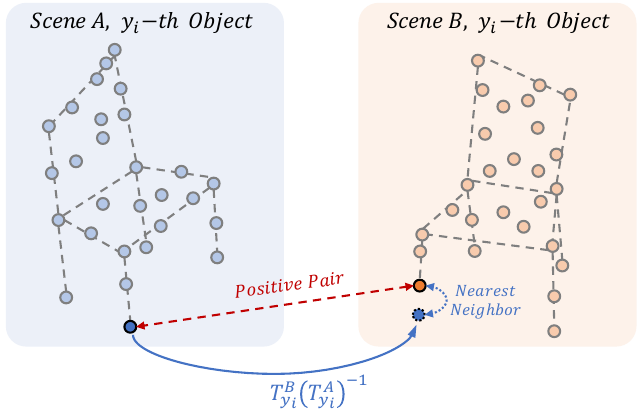}
    \caption{\small{Illustration of our relaxed point pair matching scheme. 
    }}
        \vspace{-0.1in}
    \label{Lpts}
\end{figure}

With synthetic scenes yielded by randomly placed objects, we have the access to the object transformation matrices in $\operatorname{SE}(3)$ space: $\bm{\mathcal{T}}_A=\{\bm{T}_k^A\}_{k=1}^{K},\,\bm{\mathcal{T}}_B=\{\bm{T}_k^B\}_{k=1}^{K}$, which represent the object transformations from the original ones centering at the origin of the corresponding objects in scene $A$ and scene $B$, respectively. Formally, we denote the homogeneous coordinates of $n$ seed points sampled by PointNet++ as $\bm{\mathcal{P}}^\mathcal{X}\in\mathbb{R}^{n\times 4}$. Note that $\bm{\mathcal{P}}^\mathcal{X}=\{\bm{x}_k^\mathcal{X}\}_{k=1}^{K}$, 
where $\bm{x}_k^\mathcal{X}$ represents the set of $k$-th object points. We first leverage Farthest Point Sampling (FPS) on $\bm{\mathcal{P}}^\mathcal{X}$ to sample $M$ foreground points with indices $\{m_i\}_{i=1}^M$. The corresponding point coordinates are $\{\bm{\mathcal{P}}_{m_i}^A\}_{i=1}^M$ with object correspondences $\{y_i\}_{i=1}^M$, where $y_i\in\{1,2,...,{K}\}$ represents the corresponding object index of $\bm{\mathcal{P}}_{m_i}^A$. 
Subsequently, we geometrically translate the sampled points to scene $B$, \textit{i.e.} for each point $\bm{\mathcal{P}}_{m_i}^A$ we perform:
\vspace{-0.05in}
\begin{equation}
    \Tilde{\bm{\mathcal{P}}}_{m_i}^B = \bm{T}_{y_i}^B(\bm{T}_{y_i}^A)^{-1}\bm{\mathcal{P}}_{m_i}^A,
\end{equation}
where $\Tilde{\bm{\mathcal{P}}}_{m_i}^B$ should represent the coordinate on the corresponding part of the same object in scene $B$. Then we calculate the distances from the points $\bm{x}_{y_i}^B$ to $\Tilde{\bm{\mathcal{P}}}_{m_i}^B$ and match the closest point in $\bm{x}_{y_i}^B$ as the positive point pair of $\bm{\mathcal{P}}_{m_i}^A$, as shown in Figure \ref{Lpts}. Denoting the seed point indices of $y_i$-th object as $\bm{w}_{y_i}=\{t\,|\,\bm{\mathcal{P}}_t^B\in\bm{x}_{y_i}^B\}$, the index of the matched point in scene $B$ is obtained by: 
\vspace{-0.05in}
\begin{equation}
    j^* = \underset{j\in\bm{w}_{y_i}}{\arg\min}\ \mathcal{D}(\bm{\mathcal{P}}_{j}^B, \Tilde{\bm{\mathcal{P}}}_{m_i}^B),
\end{equation}
where $\mathcal{D}(\cdot,\cdot)$ is the Euclidean distance. 
Note that the paired scenes are individually augmented through the occlusion simulation scheme, which results in random incompleteness among the paired point clouds. Therefore, we set a threshold $\theta$ on $d_i= \mathcal{D}(\Tilde{\bm{\mathcal{P}}}_{m_i}^B, \bm{\mathcal{P}}_{j^*}^B)$ to mask out the point pairs where the counterpart in scene $B$ is removed due to occlusion, and obtain the successfully matched seed point indices as $\mathcal{M} = \{(m_i, j^*)\,|\,d_i<\theta\}$, where we set $\theta=0.1$ in the experiments. With the matched point pairs, the object-aware point-level contrastive loss is defined as follow:


\scalebox{0.8}{\parbox{.9\linewidth}
{
\begin{align}
\centering
    \mathcal{L}_{pts} = & -\frac{1}{|\mathcal{M}|}\sum_{(i,j)\in\mathcal{M}}\text{log}\,\frac{\text{exp}(\bm{h}_i^A\cdot\bm{h}_{j}^B/\tau)}{\text{exp}(\bm{h}_i^A\cdot\bm{h}_{j}^B/\tau)+\sum_{\bm{h}\in\mathcal{H}_i}\text{exp}(\bm{h}_i^A\cdot\bm{h}/\tau)}\nonumber\vspace{0.5ex}\\
     &-\frac{1}{|\mathcal{M}|}\sum_{(i,j)\in\mathcal{M}}\text{log}\,\frac{\text{exp}\left(\bm{h}_{j}^B\cdot\bm{h}_i^A/\tau\right)}{\text{exp}(\bm{h}_{j}^B\cdot\bm{h}_i^A/\tau)+\sum_{\bm{h}\in\mathcal{H}_i}\text{exp}(\bm{h}_{j}^B\cdot\bm{h}/\tau)},
\end{align}
}}
Here $\bm{h}_i^\mathcal{X}$ is the projected feature of point $\bm{\mathcal{P}}_i^\mathcal{X}$ as mentioned in the object-level loss (\ref{L_obj}), $\mathcal{H}_i$ is the set of projected features of other successfully paired points from \textit{different objects} within the batch
. The proposed $\mathcal{L}_{pts}$ is constructed between the point-level samples to provide fine-grained constraints, which is integrated with $\mathcal{L}_{obj}$ to form coarse-to-fine contrastive learning.

It is worth highlighting that our relaxed point pair matching benefits the pre-training through its orthogonality to individual seed point sampling on the paired scenes, and the object-awareness of the proposed loss is two-fold: matching positive point pairs based on relative object poses, and choosing negative samples from inter-object points.

\begin{table*}[htbp]
    \small
    \centering
    \setlength{\tabcolsep}{5mm}{
    \begin{tabular}{c|c|c|cc|cc}
        \toprule
         \multirow{2}{*}{}  & \multirow{2}{*}{Pretraining dataset} & \multirow{2}{*}{Encoder} & \multicolumn{2}{c|}{ScanNetV2} & \multicolumn{2}{c}{SUN RGB-D} \\
         & & &mAP$_{25}$ & mAP$_{50}$ & mAP$_{25}$ & mAP$_{50}$ \\
         \midrule

         VoteNet & \usym{2718} &SR U-Net &56.7 &35.4 & 55.6 & 31.7\\
         PointContrast \cite{pointcontrast}&ScanNetV2$^*$&SR U-Net &58.5& 38.0 & 57.5 & 34.8\\
         
         CSC \cite{scenecontrast} &ScanNetV2$^*$&SR U-Net & 59.1 & 39.3 & 58.9 & 36.4 \\

         \midrule
         VoteNet$^\dag$& \usym{2718} &PointNet++& 60.0 &36.9 & 57.8 & 33.2\\
         CSC \cite{scenecontrast}$^\dag$&ScanNetV2$^*$ &PointNet++  & 62.5 & 39.4 & 59.2 & 34.7\\
         RandomRooms \cite{randomrooms}$^\dag$ &Synthetic&PointNet++& 61.3 &38.2 & 59.3 & 35.4\\
         DPCo \cite{dpco}$^\dag$&ScanNetV2$^*$&PointNet++& 63.0 &40.2 & 60.0 & 35.8\\
         \textbf{GRL (ours)}&Synthetic&PointNet++ & \textbf{63.9} & \textbf{41.0} & \textbf{60.7} & \textbf{36.8}\\
         \bottomrule
         
    \end{tabular}
    }
    \caption{\small{\textbf{3D object detection results with full supervision comparing to the state-of-the-arts.} $^\dag$ indicates our reproduced results using the same experiment settings. $^*$ indicates extra information in addition to point clouds, \textit{e.g.} multi-view point correspondences, depth maps.
    }}
      \vspace{-0.1in}
    \label{compare_ScanNetV2}
\end{table*}

\begin{table*}[htbp]
    \small
    \centering
    {
    \begin{tabular}{c|cc|cc|cc|cc|cc}
        \toprule
         \multirow{2}{*}{Method} & \multicolumn{2}{c|}{$\alpha=100\%$} & \multicolumn{2}{c|}{$\alpha=50\%$} & \multicolumn{2}{c|}{$\alpha=25\%$} & \multicolumn{2}{c|}{$\alpha=10\%$} & \multicolumn{2}{c}{$\alpha=5\%$} \\
         &mAP$_{25}$ & mAP$_{50}$&mAP$_{25}$ & mAP$_{50}$&mAP$_{25}$ & mAP$_{50}$&mAP$_{25}$ & mAP$_{50}$&mAP$_{25}$ & mAP$_{50}$ \\
         \midrule
         VoteNet Baseline \cite{votenet}&60.0 &36.9 & 54.0 & 30.2 & 46.0 & 23.6 & 35.1 & 15.0 & 18.7 & 5.1 \\
         CSC \cite{partition} & 61.4 &38.4 & 56.3 & 32.8 & 50.7 & 27.5 & 38.1 & 15.0 & 22.1 & 5.4 \\
         RandomRooms \cite{randomrooms}  & 61.3 &38.2 & 55.9 & 32.9 & 48.9 & 25.8 & 37.7 & 16.9 & 25.8 & 7.0\\
         DPCo \cite{dpco} & 63.0 & 40.2 & 56.1 & 33.8 & 49.5 & 27.0 & 39.3 & 19.8 & 27.3 & 10.4 \\
         \textbf{GRL (ours)} & \textbf{63.9} & \textbf{41.0} & \textbf{57.8} & \textbf{34.2} & \textbf{51.7} & \textbf{28.9} & \textbf{42.8} & \textbf{21.7} & \textbf{30.5} & \textbf{12.4}\\
         \midrule
         
    
        \midrule
         \multirow{2}{*}{Method}  & \multicolumn{2}{c|}{$all$} & \multicolumn{2}{c|}{$m=7$} & \multicolumn{2}{c|}{$m=4$} & \multicolumn{2}{c|}{$m=2$} & \multicolumn{2}{c}{$m=1$} \\
         &mAP$_{25}$ & mAP$_{50}$&mAP$_{25}$ & mAP$_{50}$&mAP$_{25}$ & mAP$_{50}$&mAP$_{25}$ & mAP$_{50}$&mAP$_{25}$ & mAP$_{50}$ \\
         \midrule
         VoteNet Baseline &60.0 &36.9& 55.3 & 31.9 & 47.4 & 26.7 & 40.9 & 16.8 & 25.5 &6.2 \\
         CSC  & 61.4 &38.4& 56.2 & 33.9 & 48.6 & 26.4 & 42.0 & 18.8  & 22.2 & 5.9\\
         RandomRooms   & 61.3 &38.2& 54.8 & 33.2 & 48.1 & 26.4 & 40.3 & 19.3  & 27.0 & \textbf{8.5}\\
         DPCo & 63.0 & 40.2 & 57.2 & 35.0 & 49.3 & 27.7 & 41.9 & 19.5 & 25.8 & 7.5\\
         \textbf{GRL (ours)} & \textbf{63.9} & \textbf{41.0}& \textbf{57.8} & \textbf{36.5} & \textbf{50.1} & \textbf{28.8}& \textbf{42.5} & \textbf{20.3}&\textbf{27.5} & 8.1    \\
         \bottomrule
    \end{tabular}}
        \vspace{-0.1in}
    \caption{\small{\textbf{3D object detection results on ScanNetV2 val set in LR and LA scenario.} We compare the results when using $\alpha=\{50\%, 25\%, 10\%, 5\%\}$ of training set for LR scenario and $m=\{1,2,4,7\}$ of bounding boxes per scene for LA scenario.}}
    \label{LRLA}
\end{table*}

\subsubsection{Occlusion-Aware Reconstruction}
With the access to complete synthetic scene-level point clouds and the motivation to encourage the backbone to preserve the detailed geometry information, we propose a novel pretext task of occlusion-aware reconstruction, for which we devise a decoder to reconstruct the complete scene. The visualization of our proposed decoder is illustrated in supplementary material. Specifically, given the complete point cloud $\bm{R}_\mathcal{X}$ and the occluded point cloud $\Tilde{\bm{R}}_\mathcal{X}$ of scene $\mathcal{X}\in\{A,B\}$, and the point-wise feature of seed points is $\bm{z}^\mathcal{X}=g(\Tilde{\bm{R}}_X)\in\mathbb{R}^{n\times s}$, where $s=256$ is the feature dimension. We then use a shared MLP $\varphi_c(\cdot)$ to predict the offsets of coordinates and features:
\begin{gather}
    \bm{\delta}^\mathcal{X} = \varphi_c(\{\bm{\Bar{\mathcal{P}}}^\mathcal{X}, \bm{z}^\mathcal{X}\})\,\in\mathbb{R}^{n\times (3+s)}, \\
    \bm{Y}_{coarse}^\mathcal{X} = \bm{\Bar{\mathcal{P}}}^\mathcal{X} + \bm{\delta}^\mathcal{X}[:,\,:3]\,\in\mathbb{R}^{n\times 3}, \\
    \bm{h}_{coarse}^\mathcal{X} = \{\bm{Y}_{coarse}^\mathcal{X},\,\bm{z}^\mathcal{X}+\bm{\delta}^\mathcal{X}[:,\,3:]\}\ \in\mathbb{R}^{n\times(3+s)}.
\end{gather}
Here $\{\cdot,\cdot\}$ is a concatenation operator on feature dimension, $\bm{\Bar{\mathcal{P}}}^\mathcal{X}\in\mathbb{R}^{n\times 3}$ is the inhomogeneous coordinates of $\bm{\mathcal{P}}^\mathcal{X}$, $\bm{Y}_{coarse}^\mathcal{X}$ and $\bm{h}_{coarse}^\mathcal{X}$ are coarse reconstruction and coarse feature, respectively.

    

To reconstruct the dense structure, we further devise a FoldingNet \cite{foldingnet}-like structure to upsample $\bm{Y}_{coarse}^\mathcal{X}$ to $\bm{Y}_{detail}^\mathcal{X}$. Specifically, we use a 2D grid $\bm{s}\in\mathbb{R}^{u^2\times 2}$, where we set $u=3$ and repeat it $n$ times to get $\bm{s}_{rc}\in\mathbb{R}^{u^2n\times 2}$. We repeat $\bm{Y}_{coarse}^\mathcal{X}$ and $\bm{h}_{coarse}^\mathcal{X}$ $u^2$ times to generate the repeated centers $\bm{Y}_{rc}^\mathcal{X}\in\mathbb{R}^{u^2n\times 3}$ and the repeated coarse feature $\bm{h}_{rc}^\mathcal{X}\in\mathbb{R}^{u^2n\times (3+s)}$. Subsequently, we predict the dense complete point cloud $\bm{Y}_{detail}^\mathcal{X}$ using a shared MLP $\varphi_d(\cdot)$:

\vspace{-2ex}
\begin{equation}
    \bm{Y}_{detail}^\mathcal{X} = \bm{Y}_{rc}^\mathcal{X} + \varphi_d\left(\{\bm{s}^{rc}, \bm{h}_{rc}^\mathcal{X}\right\})\,\in\mathbb{R}^{u^2n\times 3}.
\end{equation}

We downsample the complete dense point cloud $\bm{R}_\mathcal{X}$ to obtain ground truths for detailed point cloud and coarse point cloud, and form reconstruction loss $\mathcal{L}_{rec}$ as the addition of Chamfer Distance \cite{chamfer} for coarse and detail point clouds individually. Conceptually, our reconstruction pipeline operates as a point cloud Variational AutoEncoder (VAE) with occlusion awareness. This characteristic empowers the backbone to maintain rich geometrical information, subsequently enhancing the efficacy of transference results for specific downstream tasks. These tasks typically involve datasets with partially represented objects but demand exact localization, a common requirement in tasks such as 3D object detection.

\subsubsection{Training Objective}
The overall loss of GRL is given by:
\begin{equation}
    \mathcal{L}_{overall} = \mathcal{L}_{obj} + \lambda_{pts}\mathcal{L}_{pts} + \lambda_{rec}\mathcal{L}_{rec},
\end{equation}
where $\lambda_{pts},\ \lambda_{rec}$ are weights to balance pretext task losses. 
The joint objective aims to mine the rich geometries which can be transferred to a variety of downstream tasks.

\section{Experiments}

We evaluate our approach by conducting extensive experiments in both fully-supervised and data efficient scenarios, and we also provide representation quality analysis by limiting learnable parameters during fine-tuning and show detailed ablation study. Finally, we give qualitative comparisons with baselines.






\subsection{Self-supervised Pre-training}
\label{pre_detail}

We pre-train the backbone network PointNet++ on the synthetic scenes with objects from ShapeNet \cite{shapenet}. 
We use a two-layer MLP to get 128-D embedding for contrastive learning, and set batch size as 16 during pre-training. 
For a fair comparison with RandomRooms, we pre-train the model using the same number of sampled 3D objects.

\begin{table*}[ht]
    \small
    \centering
    \setlength{\tabcolsep}{3.5mm}
    {
    \begin{tabular}{c|cc|cc|cc|cc}
        \toprule
         \multirow{2}{*}{Method}& \multicolumn{2}{c|}{$\alpha=100\%$} & \multicolumn{2}{c|}{$\alpha=25\%$} & \multicolumn{2}{c|}{$\alpha=10\%$} & \multicolumn{2}{c}{$\alpha=5\%$} \\
         &mAP$_{25}$ & mAP$_{50}$&mAP$_{25}$ & mAP$_{50}$&mAP$_{25}$ & mAP$_{50}$&mAP$_{25}$ & mAP$_{50}$ \\
         \midrule
         VoteNet Baseline \cite{votenet}&57.8 & 33.2 & 45.6 & 22.0 & 36.5 & 12.9 & 26.4 & 8.0 \\
         CSC \cite{partition}& 62.5 & 39.4 & 49.9 & 24.0 & 38.8 & 14.6  & 33.6 & 9.7 \\
         RandomRooms \cite{randomrooms} & 61.3 & 38.2 & 49.0 & 23.7 & 38.7 & 15.1 & 32.9 & 10.1\\
         DPCo \cite{dpco} & 63.0 & 40.2 & 50.0 & 24.5 & 41.3 & 18.0 & 34.9 & 11.9 \\
         \textbf{GRL (ours)} & \textbf{63.9} & \textbf{41.0} & \textbf{50.5} & \textbf{25.4} & \textbf{43.0} & \textbf{19.8} & \textbf{36.0} & \textbf{13.2}\\
         \bottomrule
    \end{tabular}}
    \vspace{-0.1in}
    \caption{\small{\textbf{3D object detection results on SUN RGB-D dataset in LR scenario.} $\alpha$ indicates available training set raito.} }
    \vspace{-0.1in}
    \label{LR_sun}
\end{table*}

\begin{table*}[htbp]
    \small
    \centering
    \setlength{\tabcolsep}{2.7mm}{
    \begin{tabular}{c|cc|cc|cc|cc|cc}
        \toprule
         \multirow{3}{*}{Method} & \multicolumn{2}{c|}{S3DIS} & \multicolumn{8}{c}{ScanNetV2} \\
         \cmidrule(r){2-11}
          & \multicolumn{2}{c|}{$\alpha=100\%$} & \multicolumn{2}{c|}{$\alpha=100\%$} & \multicolumn{2}{c|}{$\alpha=25\%$} & \multicolumn{2}{c|}{$\alpha=10\%$} & \multicolumn{2}{c}{$\alpha=5\%$} \\
          & mIoU & mAcc & mIoU & mAcc & mIoU & mAcc & mIoU & mAcc & mIoU & mAcc \\
         \midrule
         PointNet++ Baseline \cite{votenet} & 60.3 & 69.9 & 55.8 & 68.2 & 49.2 & 60.7 & 43.5 & 55.4 & 35.3 & 46.3 \\
         CSC \cite{partition}  & 62.3 & 71.5 & 57.8 & 69.3 & 51.3 & 62.6 & 45.3 & 57.0 & 40.2 & 51.3\\
           RandomRooms \cite{randomrooms}  & 62.0 & 71.3 & 57.0 & 68.9 & 51.0 & 61.9 & 45.6 & 56.5 & 38.9 & 50.3\\

        DPCo \cite{dpco} & 59.6 & 69.2 & 57.3 & 68.2 & 52.0 & 62.8 & 46.0 & 57.0 & 40.0 & 50.8 \\
         \textbf{GRL (ours)} & \textbf{63.4} & \textbf{72.5} & \textbf{58.5} & \textbf{70.3} & \textbf{52.5} & \textbf{64.0} & \textbf{46.6} & \textbf{58.7} & \textbf{40.7} & \textbf{52.0}\\
         \bottomrule
         
    \end{tabular}
    }
    \vspace{-0.1in}
    \caption{\small{\textbf{3D semantic segmentation results on S3DIS Area 5 and ScanNetV2 val set.} 
    $\alpha$ indicates available training set raito.
    }}
    \vspace{-0.1in}
    \label{table_seg}
\end{table*}

\subsection{Transfer to 3D Object Detection}
\label{detect}

\noindent\textbf{Implementations Details.}
\label{detect_detail}
When fine-tuning on 3D object detection task, we use the framework of VoteNet \cite{votenet}, which also applies PointNet++ \cite{pointnet++} as the encoder. We evaluate on ScanNetV2 \cite{scannet} and SUN RGB-D \cite{sunrgbd}, using the pre-trained PointNet++ as initialization. During fine-tuning, we set batch size as 8 and use initial learning rate 1e-2 and 1e-3 for ScanNetV2 and SUB RGB-D, respectively. We perform exponential learning rate decay of 0.1 on epoch 80, 120 and 160, 
with a maximum epoch of 180. For the baselines, we conduct pre-training for RandomRooms \cite{randomrooms}, CSC \cite{partition} and DPCo \cite{dpco} using their pre-trained models, and reproduce their fine-tuning results under our settings.

\vspace{5pt} \noindent\textbf{3D Object Detection with Full Supervision.}
Table \ref{compare_ScanNetV2} reports the results comparing with other state-of-the-art pre-training approaches in terms of mAP@0.25 and mAP@0.5 on ScanNetV2 and SUN RGB-D, where we demonstrate consistent improvement over the other PointNet++ based methods. Note that to provide a fair comparison, we average 3 individual runs to reproduce each result.



\vspace{5pt} \noindent\textbf{3D Object Detection with Limited Scenes.}
We evaluate our method on one data-efficient learning setting - Limited Scene Reconstructions (LR), \emph{i.e.} we only have access to a subset of the full training set. The results on ScanNetV2 and SUN RGB-D are shown in Table \ref{LRLA} and Table \ref{LR_sun} respectively, where our improvements become more remarkable when only very few training data is available. 

    
\vspace{5pt} \noindent\textbf{3D Object Detection with Limited Annotations.}
We also evaluate our method on another data-efficient learning setting - Limited Annotations (LA), where the average number of available annotated bounding boxes per scene is limited. Our results using different numbers of bounding box labels per scene of ScanNetV2 are shown in Table \ref{LRLA}, where we consistently and clearly outperform the baselines. 

\begin{table}[t]
    \small
    \centering
        {
        \begin{tabular}{c|cc|cc|cc}
        \toprule
         \multirow{2}{*}{} & \multirow{2}{*}{GBSG} & \multirow{2}{*}{OS} & \multicolumn{2}{c|}{Detection} & \multicolumn{2}{c}{Segmentation}\\
         && &mAP$_{25}$ & mAP$_{50}$& mIoU & mAcc \\
         \midrule
         $A$& & & 59.8 & 35.7 & 62.5 & 71.3  \\
         $B$&\checkmark& & 60.4 & 36.3 & 63.2 & 72.1 \\
         $C$& &\checkmark& 60.2 & 36.4 & 62.9 & 71.8 \\
         $D$&\checkmark&\checkmark & \textbf{60.7} & \textbf{36.8} & \textbf{63.4} & \textbf{72.5} \\

         \bottomrule
         
        \end{tabular}}
        \vspace{-0.1in}
        \caption{\small{\textbf{Ablation study on scene generation schemes.}  
        The detection and semantic segmentation results are on SUN RGB-D and S3DIS, respectively. \emph{GBSG}: Generative Bayesian Network based Scene Generation; \emph{OS}: Occlusion Simulation.
        }}
        \vspace{-0.11in}
        \label{ablation_sg}
\end{table}

\begin{table}[t]
    \small
    \centering
        \scalebox{0.9}
        {
        \begin{tabular}{c|ccc|cc|cc}
        \toprule
         \multirow{2}{*}{} & \multirow{2}{*}{$\mathcal{L}_{obj}$} & \multirow{2}{*}{$\mathcal{L}_{pts}$} & \multirow{2}{*}{$\mathcal{L}_{rec}$} & \multicolumn{2}{c|}{Detection} & \multicolumn{2}{c}{Segmentation}\\
         &&& &mAP$_{25}$ & mAP$_{50}$& mIoU & mAcc \\
         \midrule
         $E$&\checkmark& & &61.8 & 38.9 & 57.5 & 69.2 \\
         $F$&\checkmark&\checkmark& & 62.4 & 39.4 & 58.3 & 70.1\\
         $G$&\checkmark& &\checkmark& 62.9 & 40.0 & 57.9 & 69.6 \\
         $H$&&\checkmark&\checkmark& 62.7 & 39.8 & 58.1 & 69.5 \\
         $I$&\checkmark&\checkmark&\checkmark & \textbf{63.9} & \textbf{41.0} & \textbf{58.5} & \textbf{70.3} \\

         \bottomrule
         
        \end{tabular}}
        \vspace{-0.1in}
        \caption{\small{\textbf{Ablation study on losses, fine-tuned on ScanNetV2.}  
        }}
        \vspace{-0.1in}
        \label{ablation}
\end{table}

\begin{table*}[ht]
    \small
    \centering
    \setlength{\tabcolsep}{4.0mm}{
    \begin{tabular}{c|cc|cc|cc|cc}
                \toprule
                 \multirow{3}{*}{PointNet++ Init}& \multicolumn{4}{c|}{3D Object Detection} & \multicolumn{4}{c}{3D Semantic Segmentation} \\
                \cmidrule(r){2-9}
                   & \multicolumn{2}{c|}{ScanNetV2} & \multicolumn{2}{c|}{SUB RGB-D} & \multicolumn{2}{c|}{ScanNetV2} & \multicolumn{2}{c}{S3DIS}\\
                 &mAP$_{25}$ & mAP$_{50}$ &mAP$_{25}$ & mAP$_{50}$ & mIoU & mAcc & mIoU & mAcc\\
                 \midrule
                 Random& 26.6 & 10.1 & 28.9 & 9.9 & 22.2 & 29.8 & 43.5 & 54.1 \\
                 CSC \cite{partition} & 43.3 & 15.3 & 45.9 & 19.8 & \textbf{42.6} & 53.9 & 48.8 & 56.9 \\
                 RandomRooms \cite{randomrooms} & 41.3 & 16.3 & 43.6 & 19.1 & 34.8 & 46.3 & 47.9 & 60.0  \\
                 DPCo \cite{dpco} & 33.6 & 10.3 & 39.8 & 18.0 & 28.7 & 37.5 & 37.5 & 47.0 \\
                 \textbf{GRL (ours)} & \textbf{49.7} & \textbf{26.7} & \textbf{50.5} & \textbf{24.5} & 42.0 & \textbf{54.5} & \textbf{52.5} & \textbf{62.7}  \\
                 \bottomrule
    \end{tabular}
    }
    \vspace{-0.1in}
    \caption{\small{\textbf{Representation Quality Analysis on 3D object detection and 3D semantic segmentation}. 
    }}
    \vspace{-0.1in}
    \label{freeze}
\end{table*}

\begin{figure*}[htbp]
  \centering
  \begin{subfigure}{0.24\linewidth}
    \begin{minipage}[t]{1\textwidth}
    \vspace{1pt}
    \centering
    \includegraphics[width=1\textwidth]{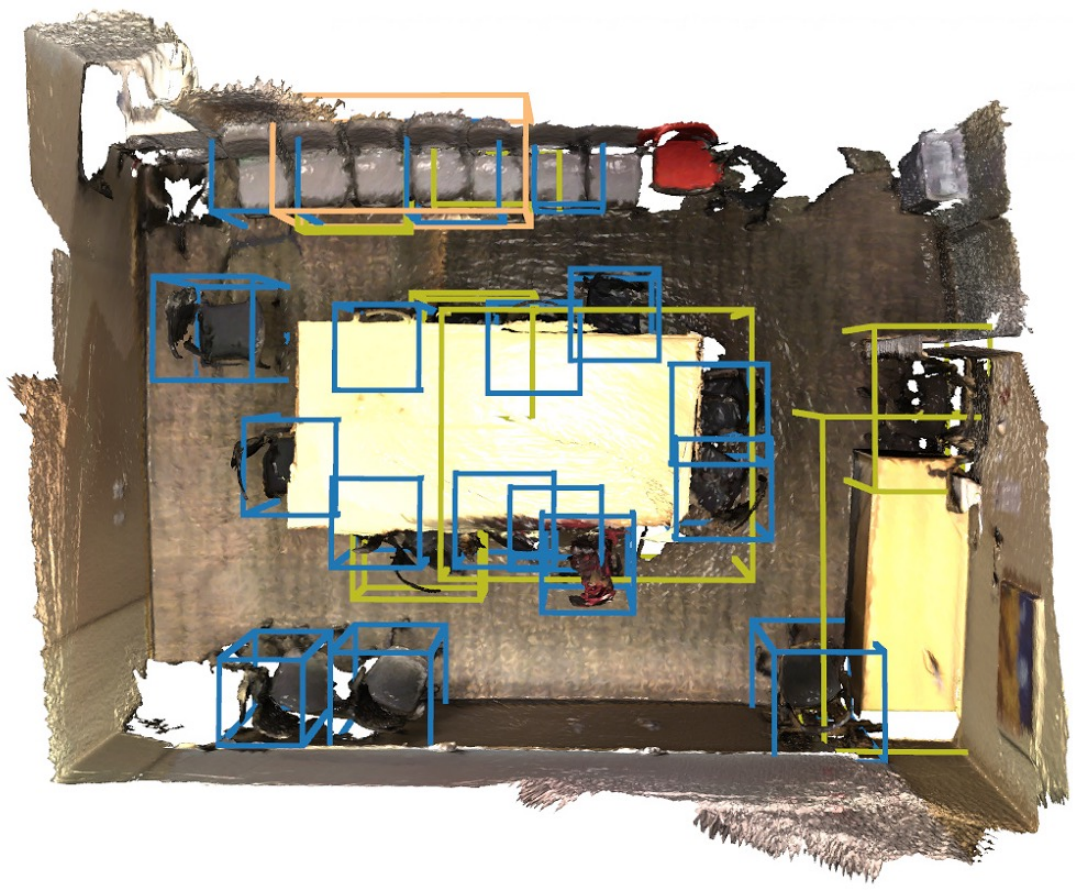}
    \end{minipage}
  \end{subfigure}
  \hfill
  \begin{subfigure}{0.24\linewidth}
     \begin{minipage}[t]{1\textwidth}
    \vspace{0pt}
    \centering
    \includegraphics[width=1\textwidth]{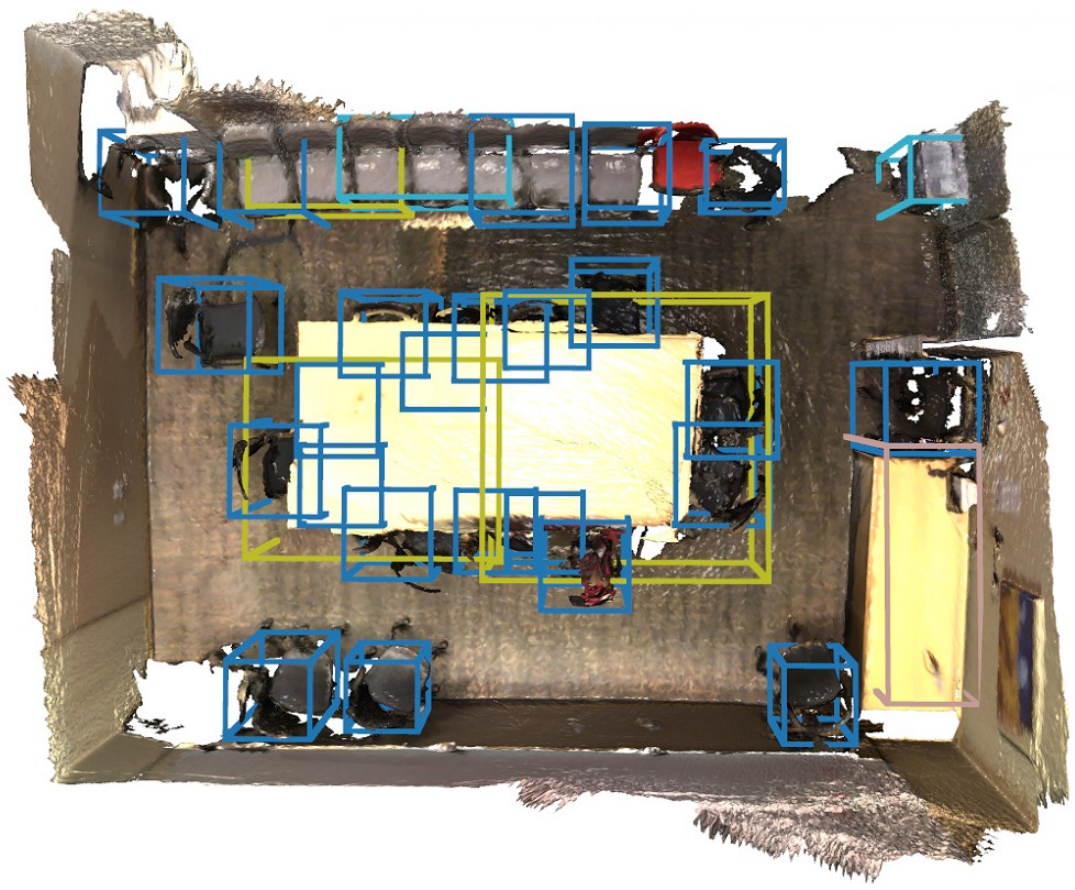}
    \end{minipage}
  \end{subfigure}
  \begin{subfigure}{0.24\linewidth}
    \begin{minipage}[t]{1\textwidth}
    \vspace{1pt}
    \centering
    \includegraphics[width=1\textwidth]{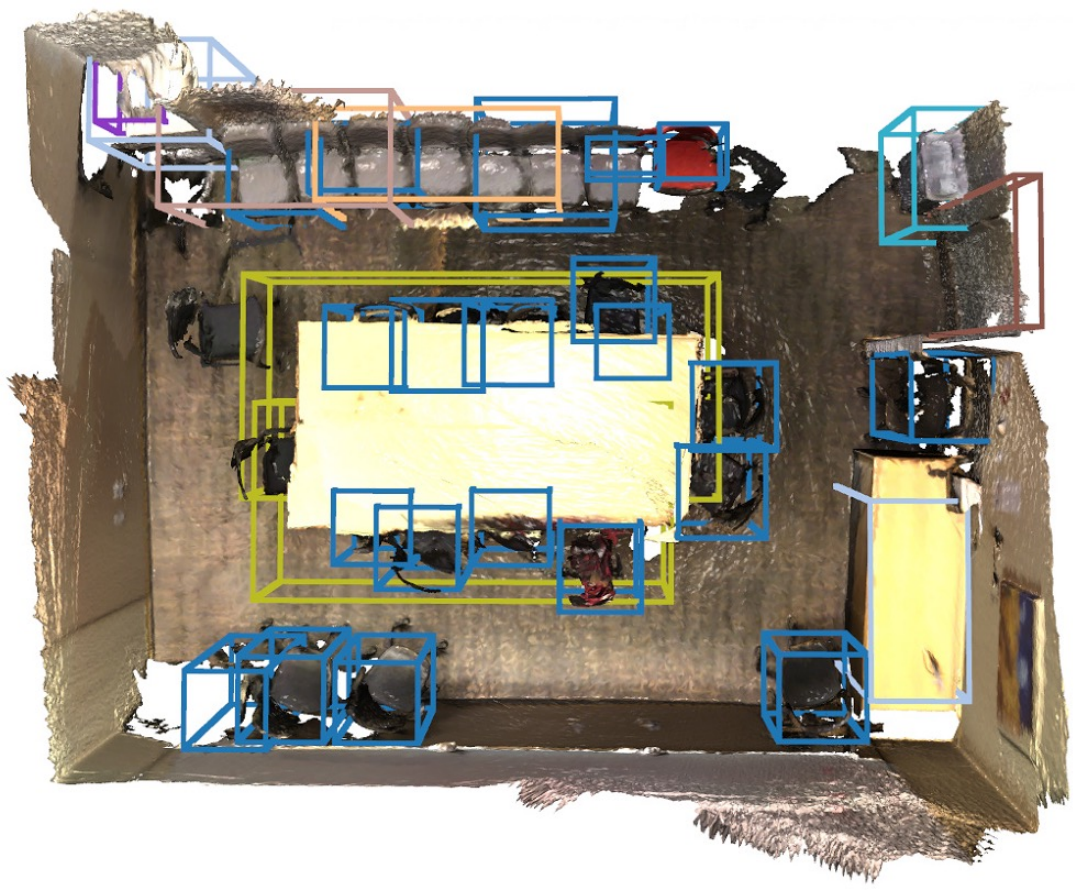}
    \end{minipage}
  \end{subfigure}
  \hfill
  \begin{subfigure}{0.24\linewidth}
    \begin{minipage}[t]{1\textwidth}
    \vspace{0pt}
    \centering
    \includegraphics[width=1\textwidth]{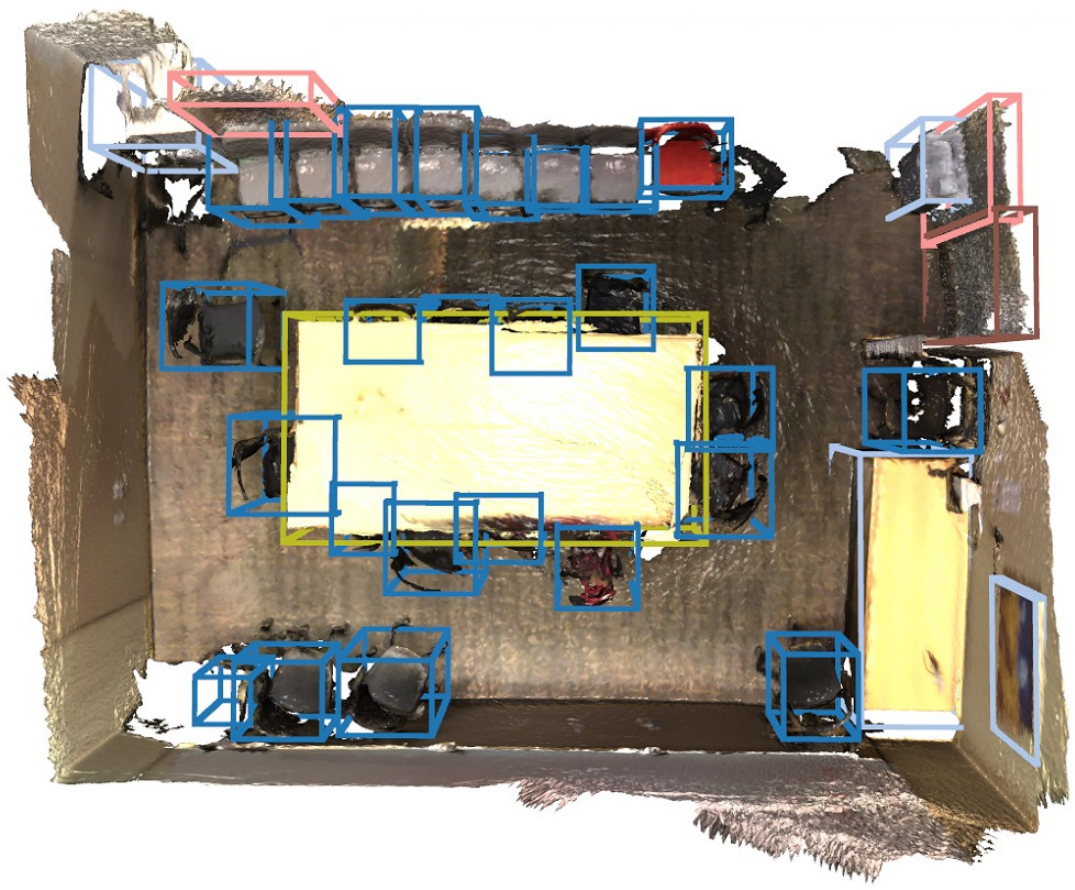}
    \end{minipage}
  \end{subfigure}
  \centerline{}
  \begin{subfigure}{0.24\linewidth}
    \begin{minipage}[t]{1\textwidth}
    \vspace{1pt}
    \centering
    \includegraphics[width=1\textwidth]{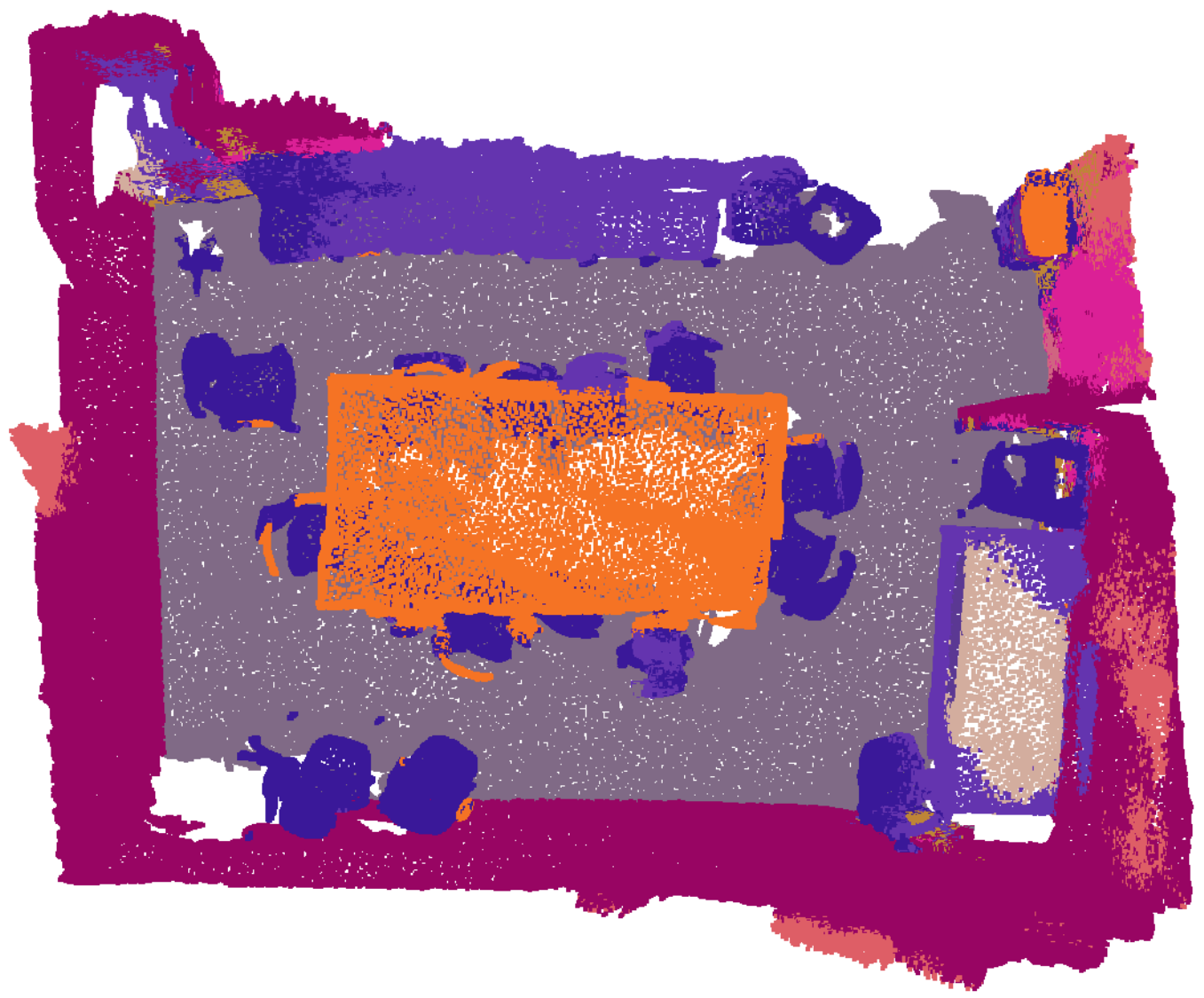}
    \end{minipage}
    \caption{VoteNet \cite{votenet}}
  \end{subfigure}
  \hfill
  \begin{subfigure}{0.24\linewidth}
     \begin{minipage}[t]{1\textwidth}
    \vspace{0pt}
    \centering
    \includegraphics[width=1\textwidth]{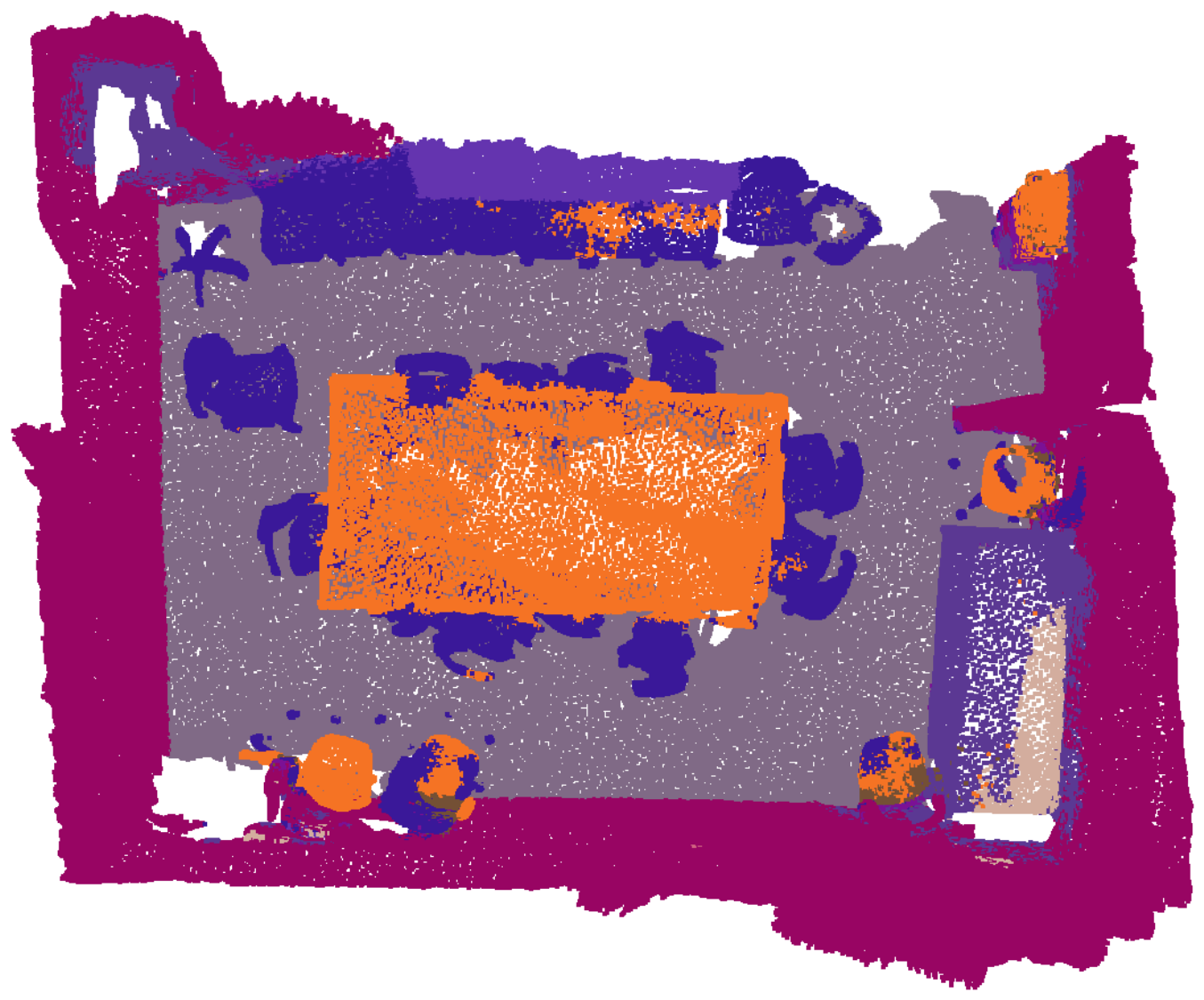}
    \end{minipage}
    \caption{DPCo \cite{dpco} + VoteNet}
  \end{subfigure}
  \begin{subfigure}{0.24\linewidth}
    \begin{minipage}[t]{1\textwidth}
    \vspace{1pt}
    \centering
    \includegraphics[width=1\textwidth]{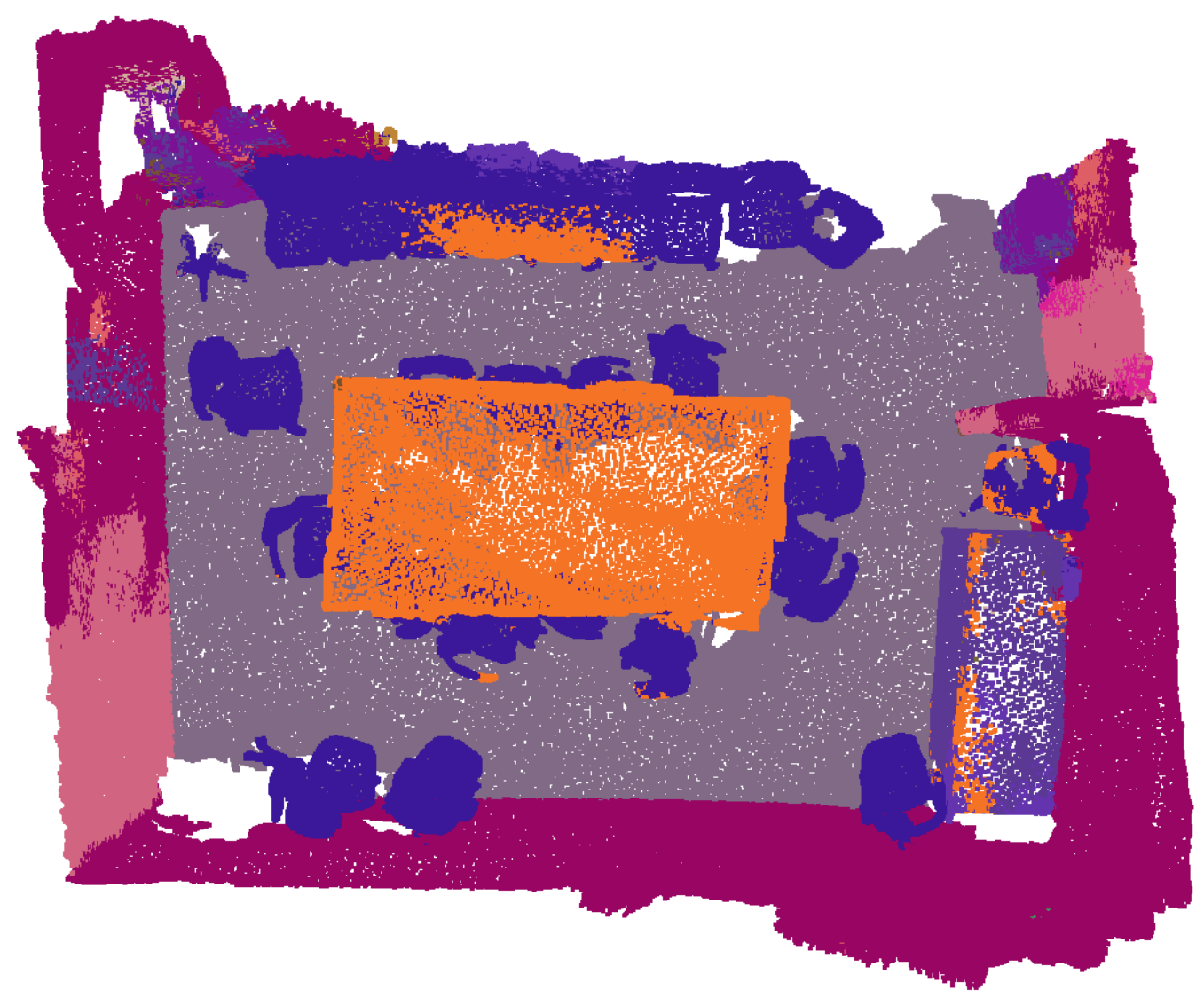}
    \end{minipage}
    \caption{Ours + VoteNet}
  \end{subfigure}
  \hfill
  \begin{subfigure}{0.24\linewidth}
    \begin{minipage}[t]{1\textwidth}
    \vspace{0pt}
    \centering
    \includegraphics[width=1\textwidth]{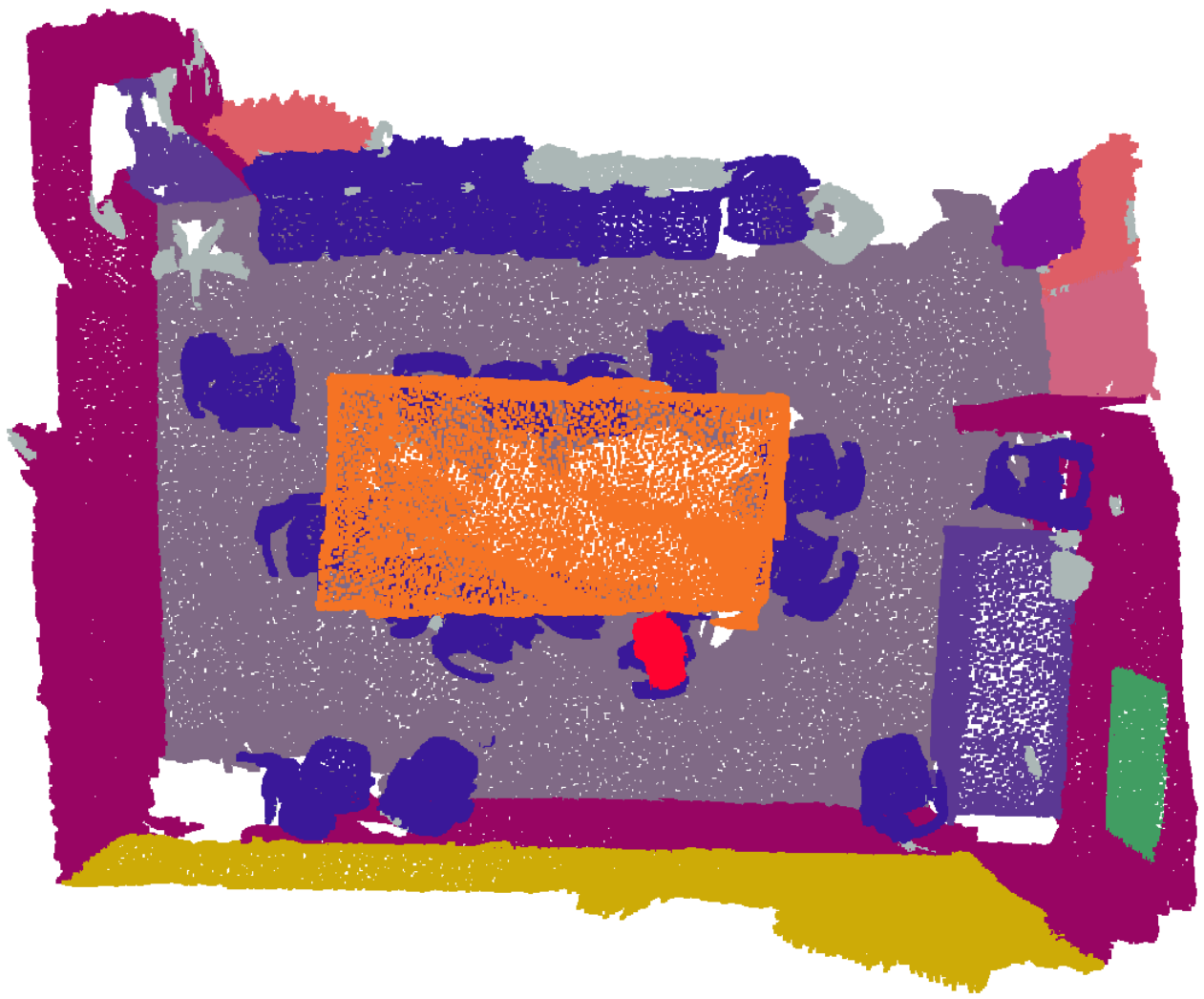}
    \end{minipage}
    \caption{Ground Truth}
  \end{subfigure}
  \vspace{-0.1in}
  \caption{\small{\textbf{Qualitative results of 3D object detection (first row) and 3D semantic segmentation (second row) when fine-tuned with 10\% of training data.} The colors of bounding boxes and segmented points represent the semantic labels. 
  Note that the two tasks are fine-tuned individually, initialized with the same pre-trained PointNet++ parameters.
  }}
  \vspace{-0.1in}
  \label{qualitative}
\end{figure*}

\subsection{Transfer to 3D Semantic Segmentation}
\label{seg}

\noindent\textbf{Implementations Details.}
To further demonstrate the generalizability of our method across different downstream tasks, we evaluate on 3D semantic segmentation. We follow the implementation of MMDetection3D \cite{mmd} and only take as input the geometry information. Specifically, we add two Feature Propagation layers after the pre-trained PointNet++ to up-sample to 4,096 points for S3DIS \cite{s3dis} and 8,192 points for ScanNetV2. We fine-tune the model for 50 epochs and 200 epochs on S3DIS and ScanNetV2, respectively, using an initial learning rate of 1e-3 with cosine decay. 

\vspace{5pt} \noindent\textbf{Results Under Different Settings.}
The results on S3DIS \cite{s3dis} and ScanNetV2 \cite{scannet} under fully-supervised and LR settings are shown in Table \ref{table_seg}, where we reach consistent improvements over RandomRooms \cite{randomrooms}, CSC \cite{partition} and DPCo \cite{dpco}. Note that although DPCo usually gives the runner-up results on other benchmarks, it instead undermines the semantic segmentation results on S3DIS, while our method gives consistently state-of-art results across benchmarks. Such improvement serves as a clear demonstration of the generalizablity of our pre-trained backbone network.

\subsection{Ablation Study}
\label{sec:ablationStudy}

\vspace{5pt} \noindent\textbf{Scene Generation.} 
In Table \ref{ablation_sg}, we ablate our scene generation strategies, where the detection and semantic segmentation fine-tuning results are reported on SUN RGB-D and S3DIS respectively, in order to demonstrate the generalizability of the scene generation schemes that are learned from ScanNet. The comparisons between rows \{$A$, $B$, $C$\} and row $D$ demonstrate the contributions of the scene generation scheme and the occlusion simulation. 

\noindent\textbf{Loss Components.} The ablation study on loss components is shown in Table \ref{ablation}. 
By only using our advanced object-level loss together with our scene generation strategies (row $E$), we give better performance comparing to Randomrooms (\cf Table \ref{compare_ScanNetV2} and Table \ref{table_seg}).
Additionally, the comparisons between rows \{$F$, $G$, $H$\} and row $I$ indicates the contribution of each loss component, illustrating the necessity of our joint-objective pre-training in extracting different perspectives of geometry information that can nourish different downstream tasks. Further ablation studies on loss components can be found in the supplementary material.

\subsection{Analytic on Representation Quality}
\label{quality}
We further explore the representation quality analysis by freezing the pre-trained parameters of PointNet++ on both downstream tasks. 
By emulating the format of linear evaluation employed in 2D representation learning \cite{bank, moco, simclr}, we create a benchmark that allows for a comparison between the quality of representations achieved using different pre-training methodologies, and concurrently reduces variability during the fine-tuning process.

The results on both tasks are shown in Table \ref{freeze}, where our approach improves with clear margins over RandomRooms \cite{randomrooms}, CSC \cite{partition} and DPCo \cite{dpco}. Note that CSC is pre-trained on ScanNetV2 with multi-view point correspondences, while we transfer the parameters pre-trained on synthetic scenes and provide comparable segmentation results on ScanNetV2, underscoring the generalizability of GRL.

\subsection{Qualitative Results}

We further provide the qualitative comparisons in Figure \ref{qualitative} when fine-tuned with 10\% data. Our model predicts more precise bounding boxes with higher recall for object detection, and higher mIoU for semantic segmentation.

\section{Conclusion}
\label{conclusion}
This paper presents GRL, a novel joint-objective representation learning approach aimed at capturing generalizable 3D geometry information. We first devise a generative Bayesian network to model the scene-object distributions when generating synthetic scenes. Subsequently, we jointly learn a coarse-to-fine contrastive learning task and an occlusion-aware reconstruction task, which enpowers the backbone to develop geometry-aware representations with broad applicability. Extensive results under various experiments settings robustly showcase the superior generalizability of GRL over other state-of-the-art pre-training methods.

\section{Acknowledge}
This work is fully done at NUS and it is supported by the Agency for Science, Technology and Research (A*STAR) under its MTC Programmatic Funds (Grant No. M23L7b0021).

{\small
\bibliographystyle{ieeenat_fullname}
\bibliography{main}
}

\end{document}


\maketitle

\appendix

This supplementary material contains the object and scene category distributions of ScanNetV2, the detailed architecture of the occlusion-aware reconstruction decoder, the implementation details, further ablation study, and a number of qualitative results.


\section{Object and Scene Category Distributions}
\label{dis}

The major object and scene category distributions of the real-world dataset ScanNet are shown in Figure \ref{obj_distri} and \ref{room_pie}, respectively.

\begin{figure}[htbp]
    \centering

\begin{tikzpicture}

\begin{axis}
[
    ybar, 
    xticklabel style={font=\tiny,
        rotate=45,anchor=mid east
    },
    xlabel={}, 
    symbolic x coords={chair,cabinet,trash can,table,pillow,sofa,lamp,bed,bag,bookshelf,computer,video display,mug,telephone,bathtub,microwave,laptop,printer,stove,bench,clock,basket,dishwasher,loudspeaker,washer,piano,mailbox,guitar,bowl}, 
    xtick=data, 
    bar width = 6pt,
    enlarge y limits = 0.05,
    enlarge x limits = 0.05,
    width = 0.5\textwidth,
    height = 7cm,
    ]
\addplot coordinates {
(chair, 4848)
(cabinet, 1798)
(trash can, 1375)
(table, 1368)
(pillow, 946)
(sofa, 503)
(lamp, 444)
(bed, 389)
(bag, 387)
(bookshelf, 253)
(computer, 246)
(video display, 220)
(mug, 166)
(telephone, 164)
(bathtub, 144)
(microwave, 141)
(laptop, 111)
(printer, 109)
(stove, 96)
(bench, 77)
(clock, 58)
(basket, 48)
(dishwasher, 43)
(loudspeaker, 43)
(washer, 42)
(piano, 39)
(mailbox, 35)
(guitar, 28)
(bowl, 24)};

\end{axis}
\end{tikzpicture}
\caption{\small{Major object category distribution of ScanNetV2.}}
\vspace{-0.1in}
\label{obj_distri}
\end{figure}

\def\defineCMYKcolor(#1,#2,#3,#4)#5{%
    \pgfmathsetmacro{\myc}{#1/4}%
    \pgfmathsetmacro{\mym}{#2/4}%
    \pgfmathsetmacro{\myy}{#3/4}%
    \pgfmathsetmacro{\myk}{#4/4}%
    \definecolor{#5}{cmyk}{\myc,\mym,\myy,\myk}%
}

\defineCMYKcolor(0,4,4,0){wedge1}
\defineCMYKcolor(0,4,3,0){wedge2}
\defineCMYKcolor(0,4,2,0){wedge3}
\defineCMYKcolor(0,4,1,0){wedge4}
\defineCMYKcolor(0,4,0,0){wedge5}
\defineCMYKcolor(1,4,0,0){wedge6}
\defineCMYKcolor(2,4,0,0){wedge7}
\defineCMYKcolor(3,4,0,0){wedge8}
\defineCMYKcolor(4,4,0,0){wedge9}
\defineCMYKcolor(4,3,0,0){wedge10}
\defineCMYKcolor(4,2,0,0){wedge11}
\defineCMYKcolor(4,1,0,0){wedge12}
\defineCMYKcolor(4,0,0,0){wedge13}
\defineCMYKcolor(4,0,1,0){wedge14}
\defineCMYKcolor(4,0,2,0){wedge15}
\defineCMYKcolor(4,0,3,0){wedge16}
\defineCMYKcolor(4,0,4,0){wedge17}
\defineCMYKcolor(3,0,4,0){wedge18}
\defineCMYKcolor(2,0,4,0){wedge19}
\defineCMYKcolor(1,0,4,0){wedge20}
\defineCMYKcolor(0,0,4,0){wedge21}
\defineCMYKcolor(0,1,4,0){wedge22}
\defineCMYKcolor(0,2,4,0){wedge23}
\defineCMYKcolor(0,3,4,0){wedge24}

\begin{figure}[htbp]
\centering
\begin{tikzpicture}

\tikzstyle{every node}=[font=\tiny]

\pie[
        rotate=50,
        radius=2.5,
        hide number,
        color = {wedge10,wedge12,wedge14,wedge16,wedge18,wedge20,wedge22,wedge24,wedge1,wedge3,wedge5,wedge6,wedge7,},
        style=drop shadow,
]{
18.445945945945947/Hotel(18.04\%),
15.135135135135135/Lounge(14.81\%),
14.324324324324325/Bathroom(14.01\%),
13.91891891891892/Room(13.62\%),
11.68918918918919/Office(11.43\%),
7.297297297297297/Kitchen(7.14\%),
4.527027027027027/Library(4.43\%),
3.6486486486486487/Lobby(3.57\%),
2.7027027027027026/Apartment(2.64\%),
2.5/Classroom(2.45\%),
2.364864864864865/Misc.(2.31\%),
2.1621621621621623/Hallway(2.12\%),
1.2837837837837838/Storage(1.26\%)
}

\end{tikzpicture}
\vspace{-0.1in}
\caption{\small{Major scene category distribution of ScanNetV2.}}
\label{room_pie}
\end{figure}

\begin{figure*}
    \centering
    \includegraphics[width=1\textwidth]{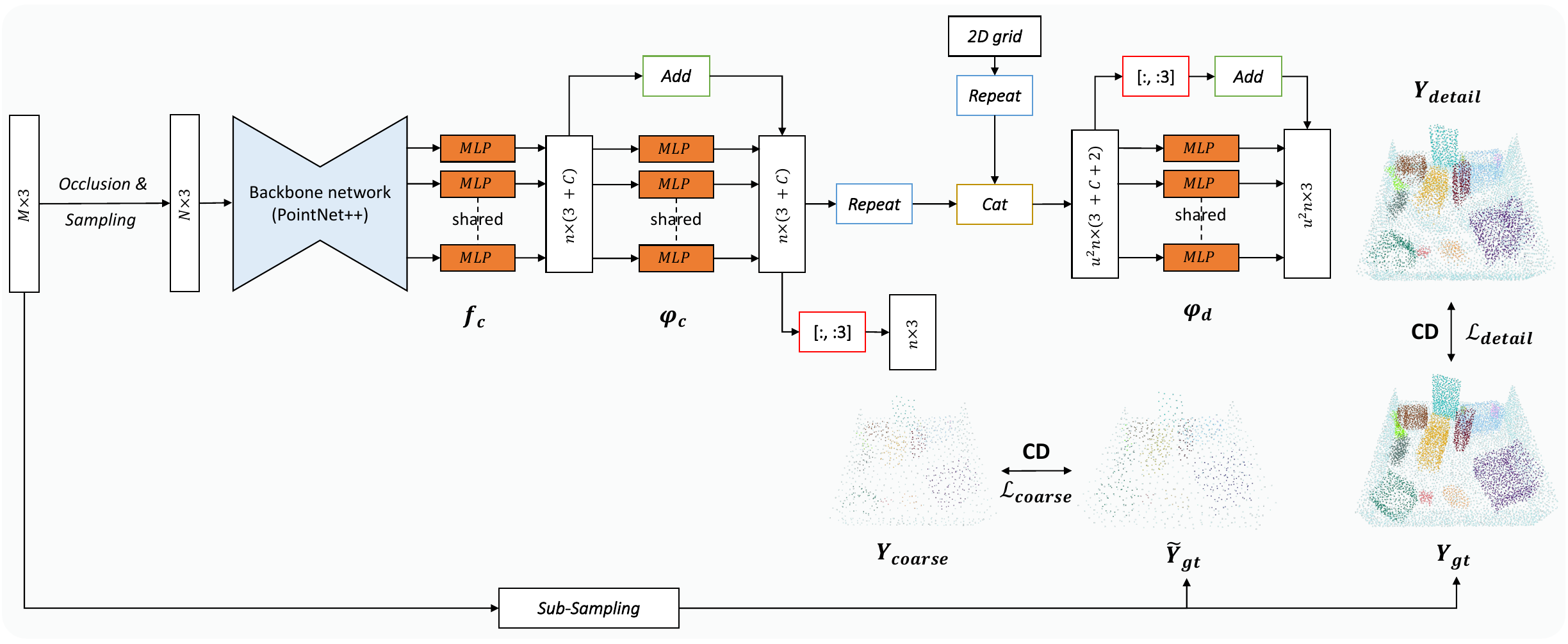}
    \caption{\small{\textbf{The Architecture of the Point Cloud Completion Decoder.} Given the point-wise features of the occluded point cloud, the decoder predicts point-wise coordinate offsets for the underlying coarse completed point cloud, which is followed by GridNet-like mapping to detailed 3D surfaces.}}
    \label{comp_decoder}
\end{figure*}

\section{Detailed Architecture of the Occlusion-aware Reconstruction Decoder}
\label{decoder}


Figure \ref{comp_decoder} illustrates the detailed architecture of our occlusion-aware reconstruction decoder. Note that instead of directly predicting the coordinates of the coarse completed point cloud based on the global feature, we predict the point-wise offsets for coordinate and feature to better reconstruct the detailed structures of scene-level point clouds.

\section{Implementation Details}

\vspace{5pt}
\noindent\textbf{Computational Cost.}\label{cost} The pre-training and fine-tuning experiments are conducted on single NVIDIA RTX 4090, which take about 2 days and 1 day, respectively.

\vspace{5pt}
\noindent\textbf{Hyperparameters.} For the $\epsilon$-greedy strategy in scene generator, we set $\epsilon=0.1$, \textit{i.e.} for each $x_C$ we use $p(X_C \mid X_S = x_s)$ with possibility of 0.9 and use uniform distribution with possibility of 0.1. 
In terms of the pre-training objectives, we sample $M=100$ points for the point-level contrastive loss, and temperature $\tau=0.03$ for hierarchical contrastive losses. We set $\lambda_1=0.1$, $\lambda_2=100$ in overall loss to combine different pretext-task losses. 

\vspace{5pt}
\noindent\textbf{Algorithm runs}
For the fine-tuning experiments, we run 3 times and use the average to report each result. For the representation quality analysis and ablation study where pre-trained parameters are frozen, we use single algorithm run to report each result.

\section{Further Ablation Studies}
\label{ablation}

\begin{table}[tp]
    \small
    \centering
        {
        \begin{tabular}{c|cc|cc|cc}
        \toprule
         \multirow{2}{*}{} & \multirow{2}{*}{GBSG} & \multirow{2}{*}{OS} & \multicolumn{2}{c|}{Detection} & \multicolumn{2}{c}{Segmentation}\\
         && &mAP$_{25}$ & mAP$_{50}$& mIoU & mAcc \\
         \midrule
         $A'$& & & 43.2 & 18.8 & 38.3 & 49.3  \\
         $B'$&\checkmark& & 45.0 & 20.7 & 40.8 & 52.7 \\
         $C'$& &\checkmark&47.3 & 23.7 & 38.5 & 50.2 \\
         $D'$&\checkmark&\checkmark & \textbf{49.7} & \textbf{26.7} & \textbf{42.0} & \textbf{54.5} \\

         \bottomrule
         
        \end{tabular}}
        \caption{\textbf{Ablation study on scene generation comparing representation quality.}  
        \emph{GBSG}: Generative Bayesian network based Scene Generation; \emph{OS}: Occlusion Simulation. }
        \label{ablation_sg}
\end{table}

\begin{table}[tp]
    \small
    \centering
        \scalebox{0.9}
        {
        \begin{tabular}{c|ccc|cc|cc}
        \toprule
         \multirow{2}{*}{} & \multirow{2}{*}{$\mathcal{L}_{obj}$} & \multirow{2}{*}{$\mathcal{L}_{pts}$} & \multirow{2}{*}{$\mathcal{L}_{comp}$} & \multicolumn{2}{c|}{Detection} & \multicolumn{2}{c}{Segmentation}\\
         &&& &mAP$_{25}$ & mAP$_{50}$& mIoU & mAcc \\
         \midrule
         $E'$&\checkmark& & &43.6 & 19.2 & 38.8 & 49.2 \\
         $F'$&\checkmark&\checkmark& & 44.5 & 20.4 & 41.8 & 54.1 \\
         $G'$&\checkmark& &\checkmark& 46.3 & 22.0 & 39.2 & 50.7 \\
         $H'$&&\checkmark&\checkmark& 42.6 & 21.9 & 41.0 & 52.7 \\
         $I'$&\checkmark&\checkmark&\checkmark & \textbf{48.5} & \textbf{24.6} & \textbf{42.0} & \textbf{54.5} \\

         \bottomrule
         
        \end{tabular}}
        \caption{\textbf{Ablation study on loss components, we compare the results on ScanNetV2 when freezing PointNet++.}  
        $\mathcal{L}_{obj}$: object-level contrastive loss; $\mathcal{L}_{pts}$: point-level contrastive loss; $\mathcal{L}_{rec}$: occlusion-aware reconstruction loss. } \vspace{-4mm}
        \label{ablation_loss}
\end{table}

Table \ref{ablation_sg} and Table \ref{ablation_loss} show the ablation results when freezing the pre-trained PointNet++, where the margins between different methods are shown more clearly. The results demonstrate similar conclusions to the ones in the main paper. Table \ref{fix_seed} studies the effectiveness of our distance-based point matching strategy comparing to the exact matching, where the exact matching forces the same object point seeds in a pair of scenes and undermines the robustness of the pre-trained model to the sampling randomness.

\begin{table}[tp]
    \small
    \centering
        {
        \begin{tabular}{c|cc|cc}
        \toprule
         \multirow{2}{*}{} & \multicolumn{2}{c|}{Detection} & \multicolumn{2}{c}{Segmentation}\\
         &mAP$_{25}$ & mAP$_{50}$& mIoU & mAcc \\
         \midrule
         Exact&45.3 & 22.2 & 39.3 & 50.9 \\
         \textbf{Ours}& \textbf{49.7} & \textbf{26.7} & \textbf{42.0} & \textbf{54.5} \\

         \bottomrule
         
        \end{tabular}}
        \caption{\textbf{Ablation study on point matching strategies evaluating representation quality on ScanNetV2.} Exact: we alternatively conduct pre-training where we use the same point seeds such that we have exact point correspondences.} 
        \label{fix_seed}
\end{table}

\begin{table}[htbp]
    \footnotesize
    \centering
    \setlength{\tabcolsep}{0.9mm}
    {
    \begin{tabular}{cc|cc|cc}
        \toprule
         Dataset & SSL loss&mAP$_{25}$ & mAP$_{25}^*$&mIoU & mIoU$^*$ \\
         \midrule
         3D-FRONT & Ours$^\ddag$ & 62.5 & 42.8 & 57.4 & 32.9 \\
         GBSG (ours)$^\dag$ & PointInfoNCE loss & 61.6 & 38.2 & 57.0 & 34.9\\
         GBSG (ours) & Ours & \textbf{63.9} & \textbf{49.7} & \textbf{58.5} & \textbf{42.0}\\
         \bottomrule
    \end{tabular}}
    \caption{\textbf{Analysis on pre-training dataset and SSL loss.} $^\dag$ and $^\ddag$ indicate modifications to our GBSG and losses, respectively, to perform pre-training on a pair of identical scenes with random augmentations. $^*$ are results with frozen encoder.} 
    \label{3dfront}
\end{table}

\section{Necessity of our data generation scheme}
It is important to note that the objective of our data generation scheme is not to synthesize rooms that closely resemble those in real world. Instead, we aim to generate diverse synthetic scenes with real-world patterns, allowing our model to benefit from this diversity and the inclusion of prevalent real-world patterns. Consequently, our dataset generation scheme is intentionally designed for representation learning and diverges from the objectives of state-of-the-art scene generation methods. In Table \ref{3dfront} we compare using different pre-training dataset and SSL losses, which highlights the significance of our proposed GRL framework.

\section{Qualitative Results}
\label{qualitative}

In this section, we show different qualitative results of the object-aware point matching, the occlusion-aware reconstruction results, and the object detection results on SUN RGB-D.

\begin{figure*}
    \centering
    \includegraphics[width=0.9\linewidth]{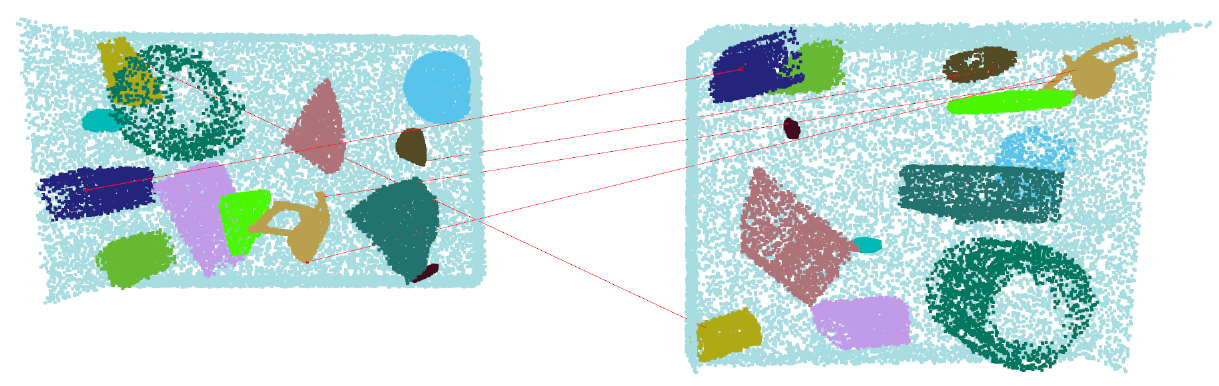}
    \caption{\small{\textbf{Qualitative results of object-aware point matching.} The matched points are linked by red lines. We show 5 point correspondences in this pair of scenes.}}
    \label{matching}
\end{figure*}

\begin{figure*}
\centering
\begin{minipage}[t]{0.32\textwidth}
\centering
\caption*{$\bm{\mathcal{P}}$}
\includegraphics[width=5cm]{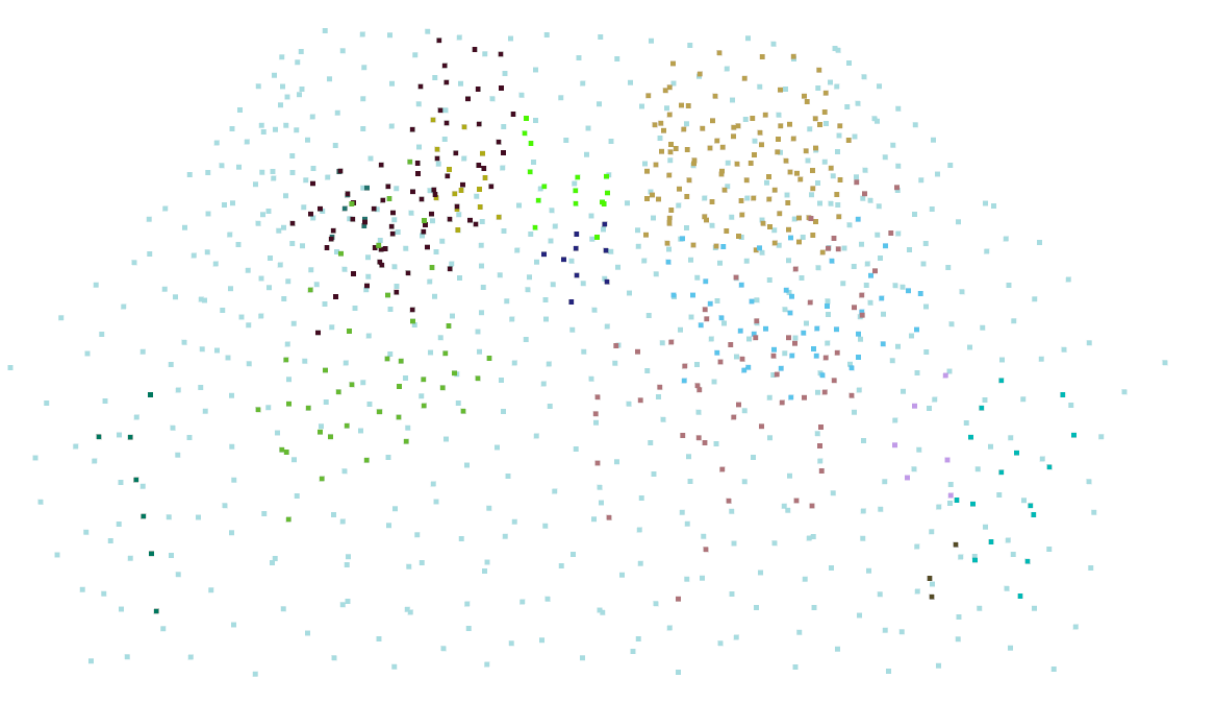}
\end{minipage}
\begin{minipage}[t]{0.32\textwidth}
\centering
\caption*{$\bm{Y}_{detail}$}
\includegraphics[width=5cm]{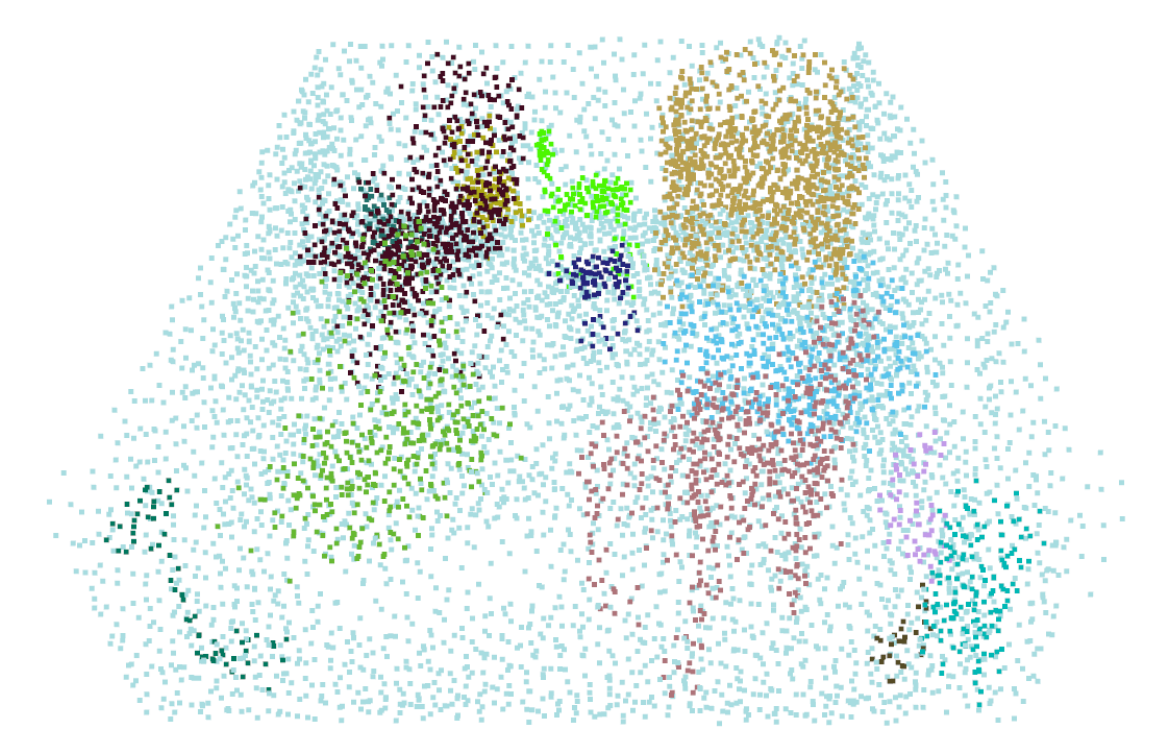}
\end{minipage}
\begin{minipage}[t]{0.32\textwidth}
\centering
\caption*{$\bm{Y}_{gt}$}
\includegraphics[width=5cm]{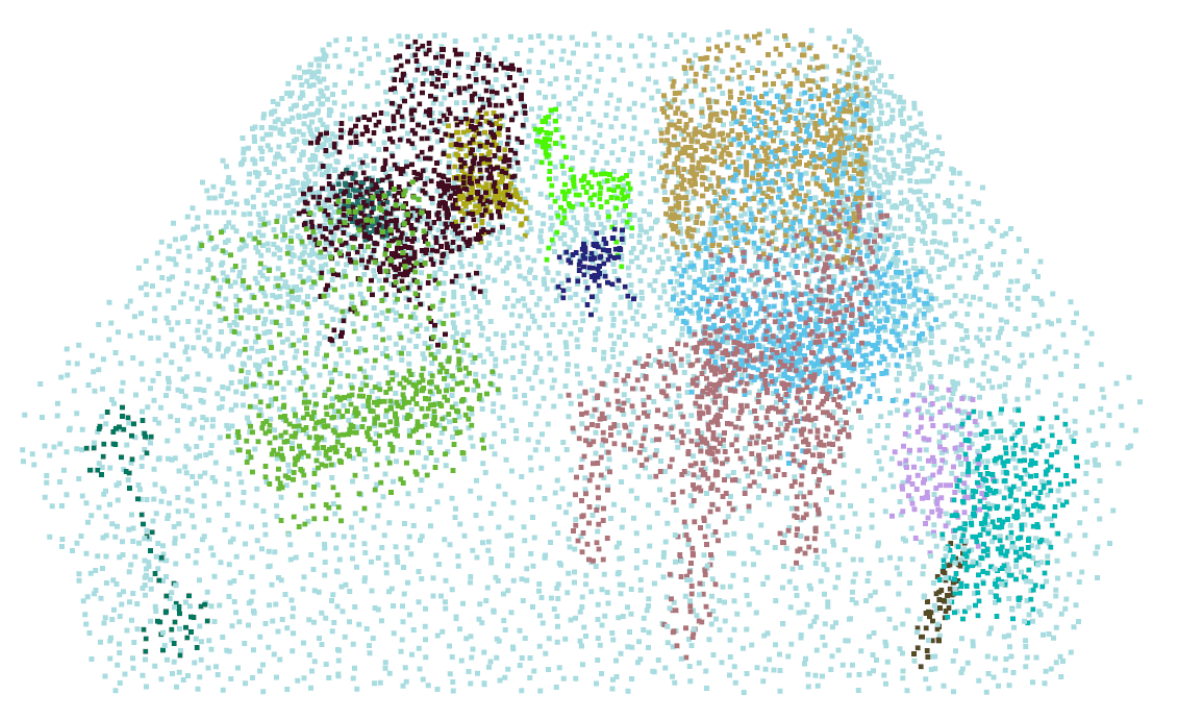}
\end{minipage}
~\\
~\\
~\\
\begin{minipage}[t]{0.32\textwidth}
\centering
\includegraphics[width=5cm]{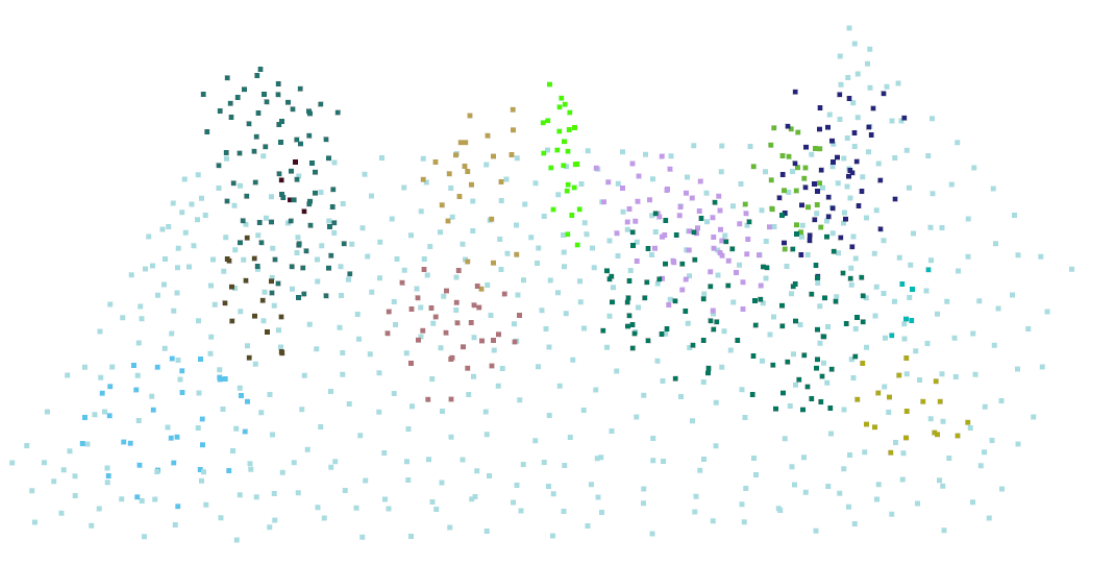}
\end{minipage}
\begin{minipage}[t]{0.32\textwidth}
\centering
\includegraphics[width=5cm]{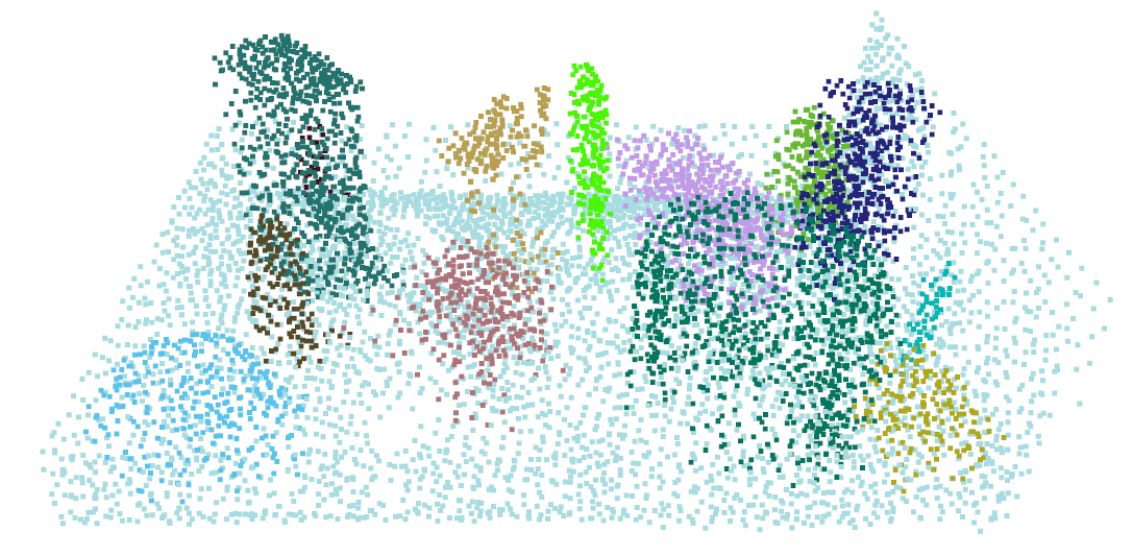}
\end{minipage}
\begin{minipage}[t]{0.32\textwidth}
\centering
\includegraphics[width=5cm]{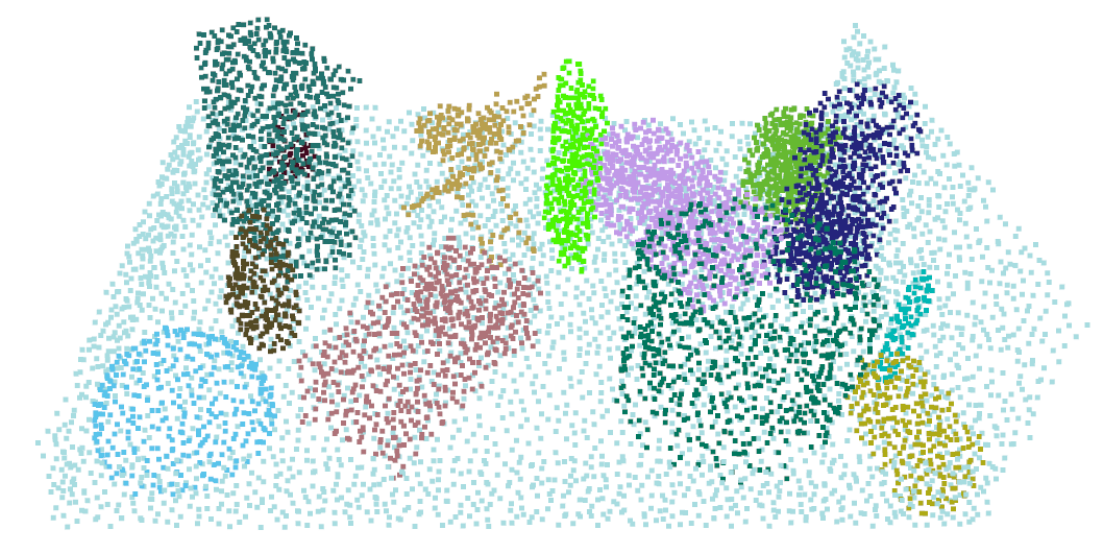}
\end{minipage}
~\\
~\\
\begin{minipage}[t]{0.32\textwidth}
\centering
\includegraphics[width=5cm]{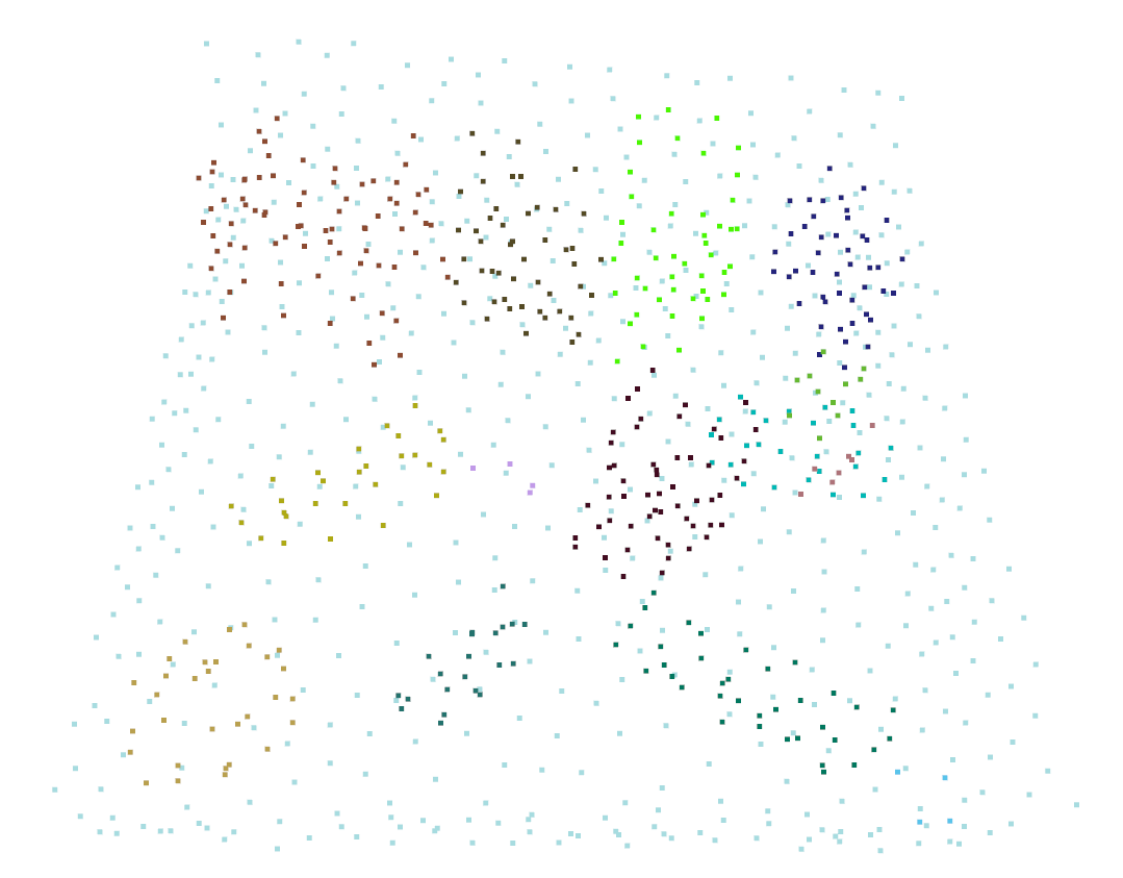}
\end{minipage}
\begin{minipage}[t]{0.32\textwidth}
\centering
\includegraphics[width=5cm]{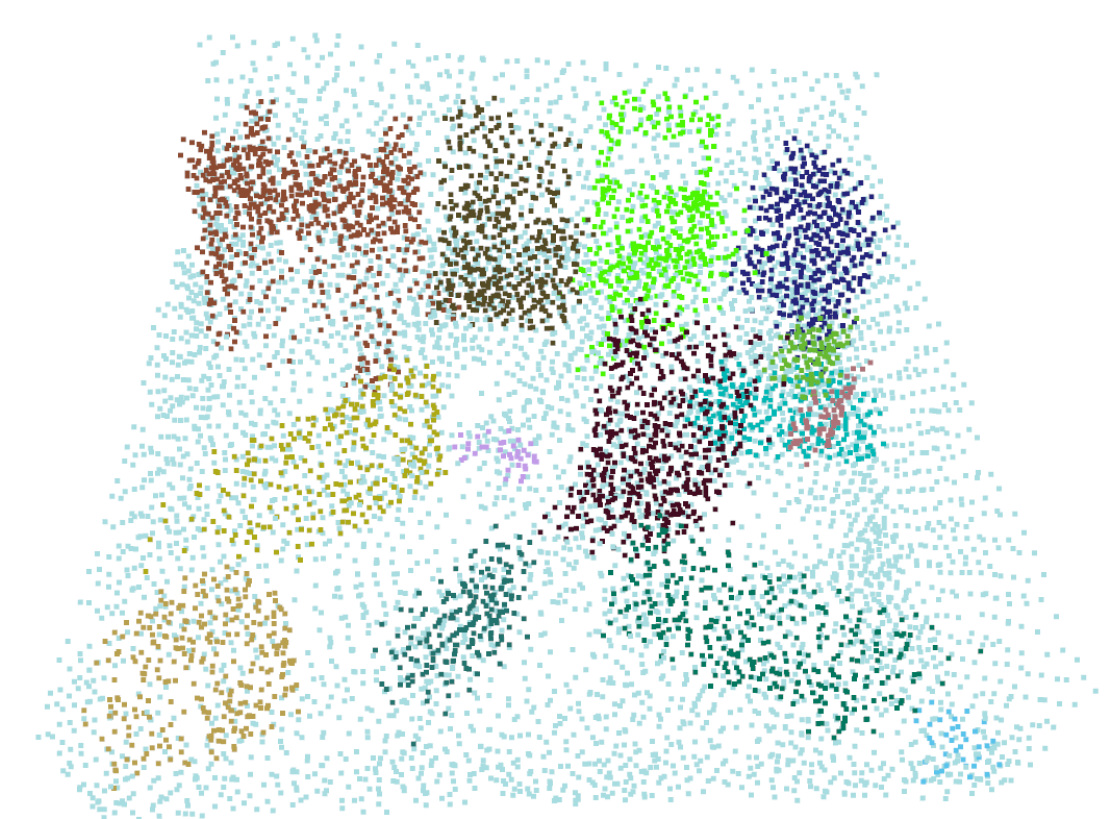}
\end{minipage}
\begin{minipage}[t]{0.32\textwidth}
\centering
\includegraphics[width=5cm]{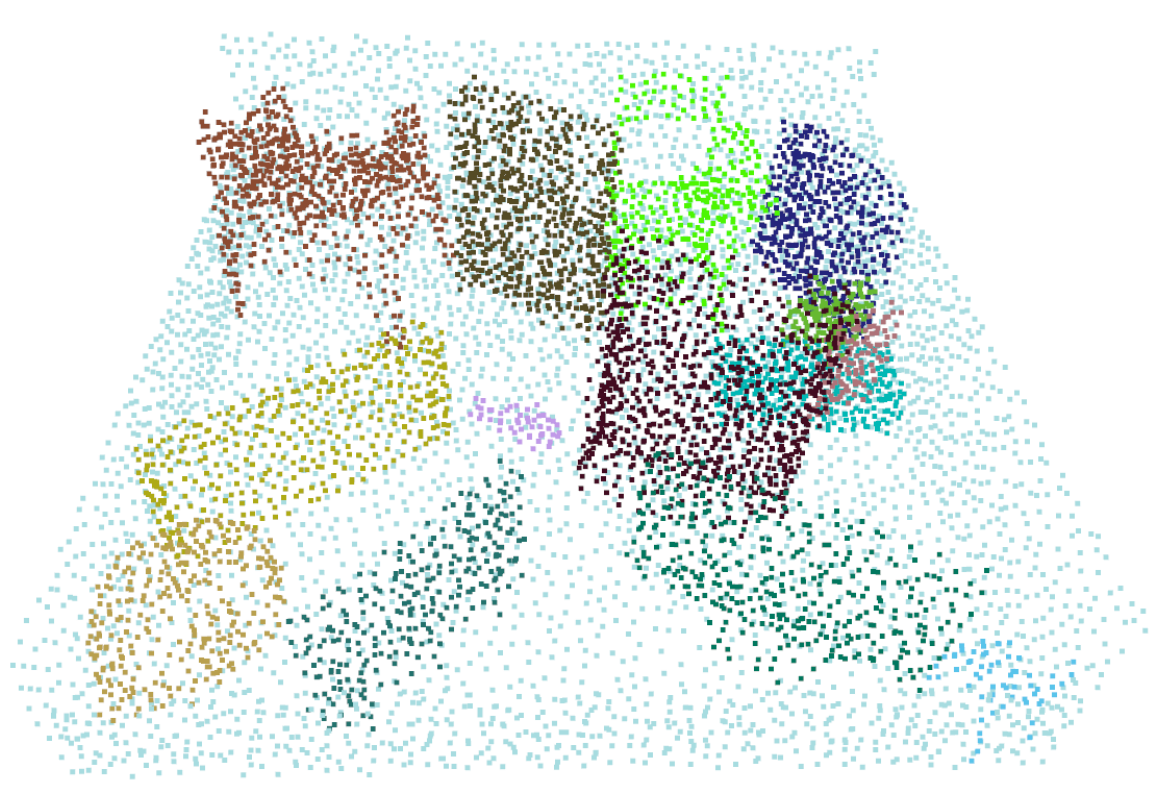}
\end{minipage}
~\\
~\\
\caption{\small{\textbf{Qualitative results of the occlusion-aware point cloud reconstruction results.} From the left to the right column are the input partial point cloud $\bm{p}$, reconstructed point cloud $\bm{Y}_{detail}$, and the corresponding ground truth $\bm{Y}_{gt}$.}}
\label{completion}
\vspace{-0.1in}
\end{figure*}

\subsection{Object-aware Point Matching}
\label{pair}
We show the positive pair matching results in Figure \ref{matching}, where five resultant point pairs are linked by red lines. As can be seen, the matched point pairs are localized in the corresponding parts of the same objects. 

\subsection{Point Cloud Reconstruction}
\label{comp}
To show the intermediate reconstruction performance, we compare the inputs of the reconstruction decoder, the reconstructed point clouds, and the corresponding ground-truths in Figure \ref{completion}. 
We can see that our decoder gives upsampled point clouds with fine-grained structures.



\begin{figure*}
    \centering
    \subfloat[VoteNet]{
        \begin{minipage}[t]{0.48\textwidth}
        \centering
        \includegraphics[width=8cm]{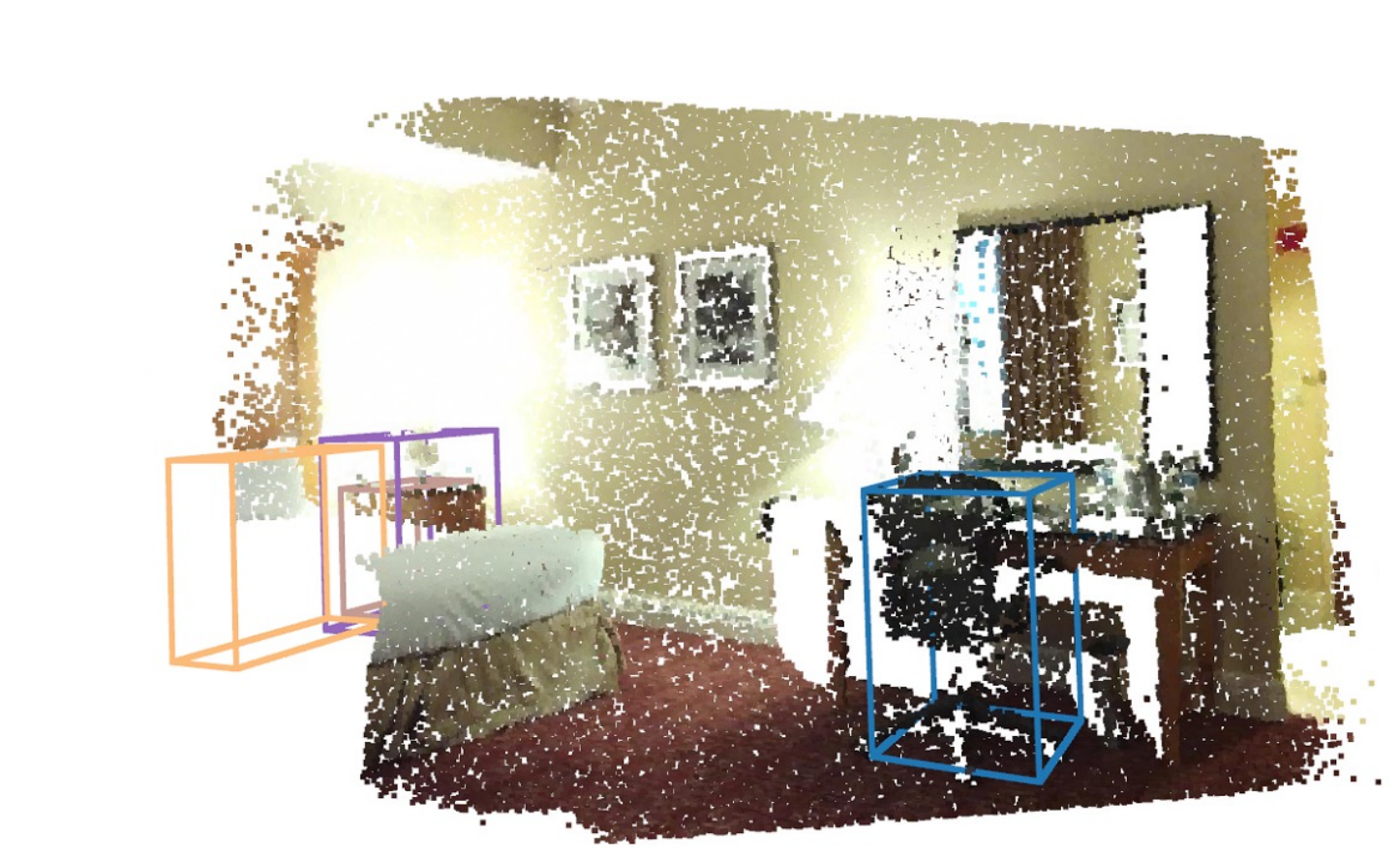}
        \end{minipage}
    }
    \subfloat[DPCo + VoteNet]{
        \begin{minipage}[t]{0.48\textwidth}
        \centering
        \includegraphics[width=8cm]{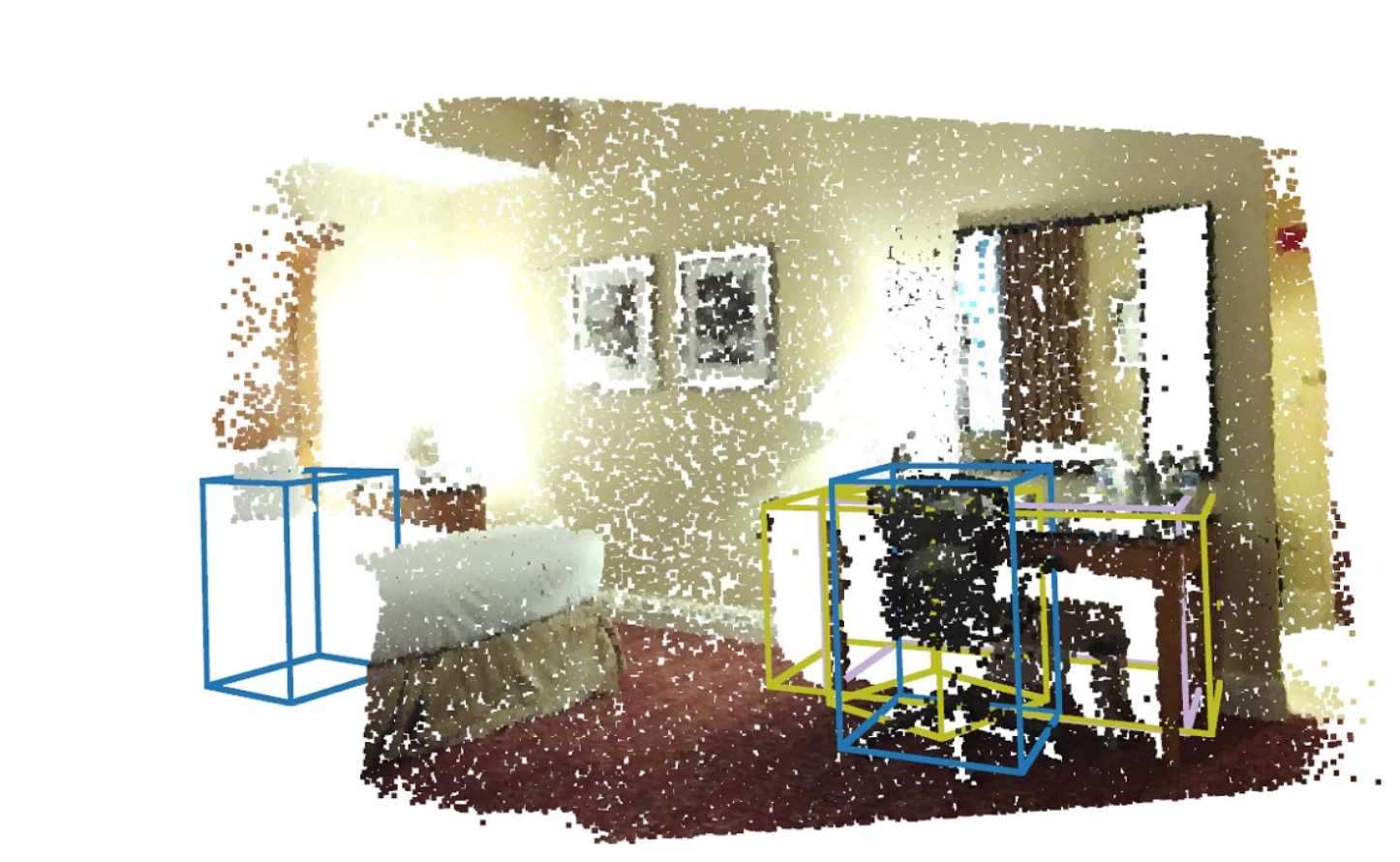}
        \end{minipage}
    }
    ~\\
    \subfloat[Ours + VoteNet]{
        \begin{minipage}[t]{0.48\textwidth}
        \centering
        \includegraphics[width=8cm]{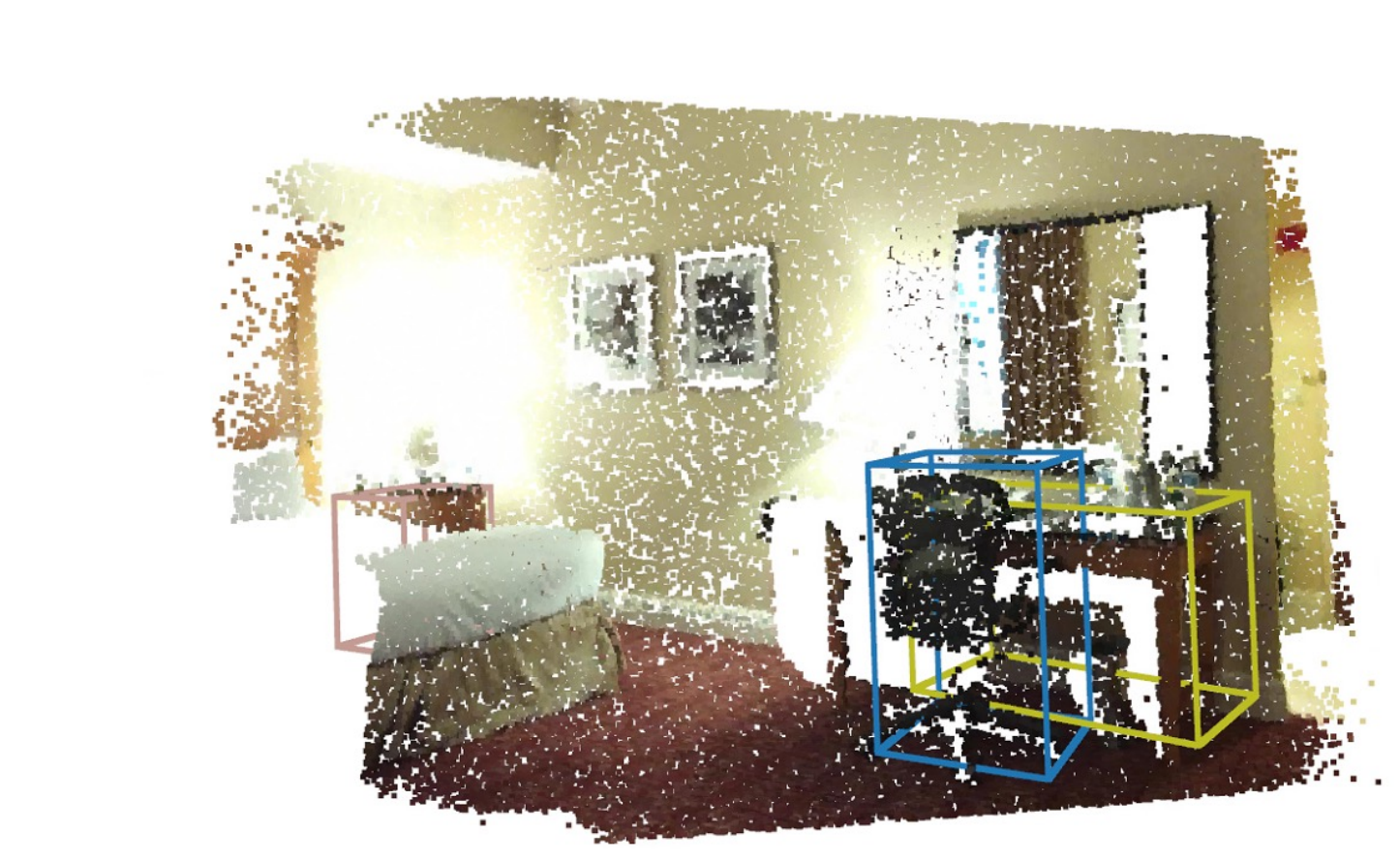}
        \end{minipage}
    }
    \subfloat[Ground Truth]{
        \begin{minipage}[t]{0.48\textwidth}
        \centering
        \includegraphics[width=8cm]{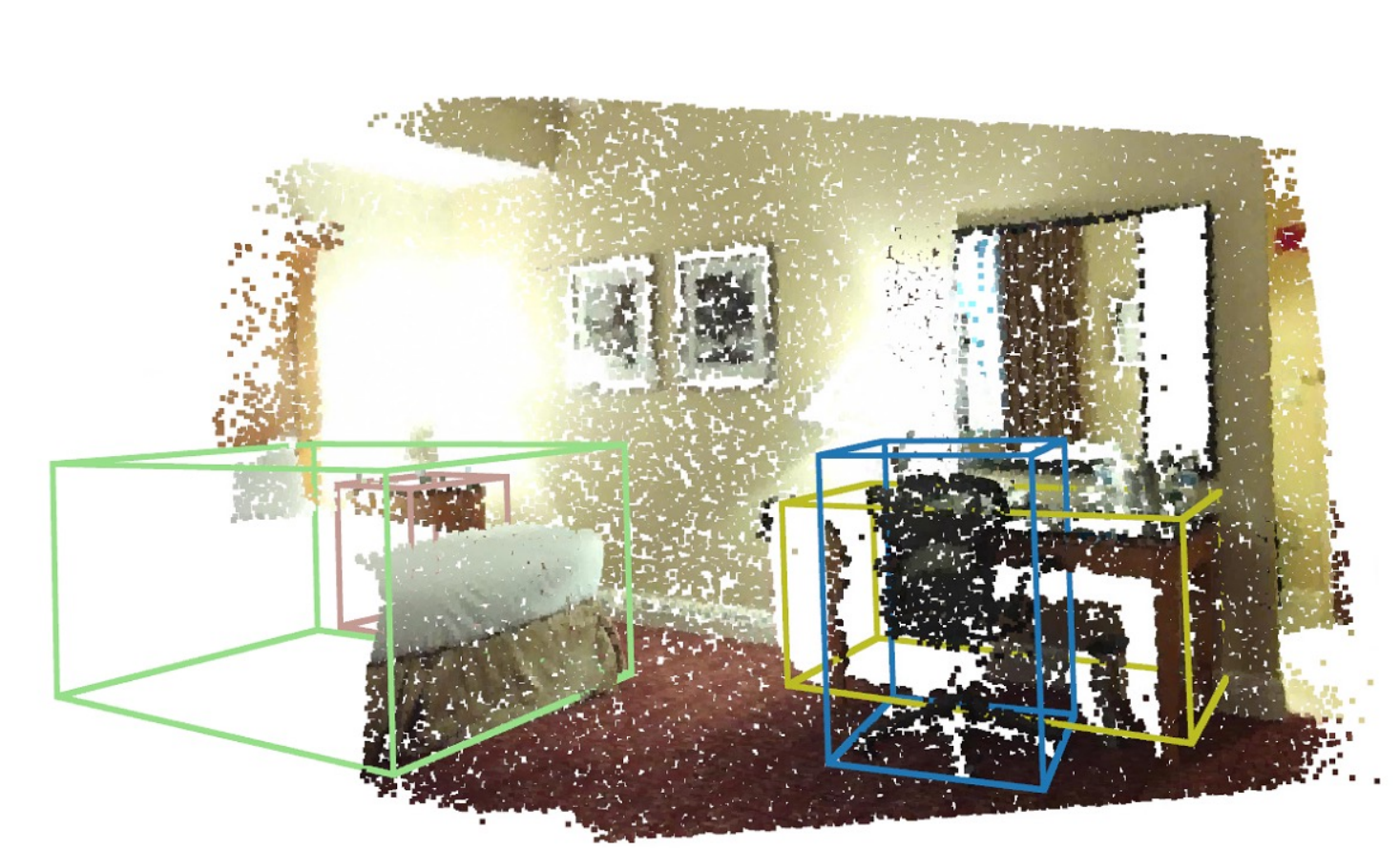}
        \end{minipage}
    }

    \caption{\small{\textbf{Qualitative results of 3D object detection on SUN RGB-D val set when fine-tuned with 10\% data.} We compare our results with baselines and ground truth when trained with 10\% labeled data. The colors of bounding boxes represent the semantic labels.}}
    \label{detect_sunrgbd}
\end{figure*}

\subsection{Object Detection on SUN RGB-D}
\label{detect}
The qualitative comparisons of 3D object detection on SUN RGB-D val set when fine-tuned with 10\% data are shown in Figure \ref{detect_sunrgbd}, respectively. Comparing to the baselines, our method gives more precise bounding boxes localization with higher recall.